%% file: main.tex
\documentclass[lettersize,journal]{IEEEtran}
\usepackage{amsmath,amsfonts}
\usepackage{algorithmic}
\usepackage{algorithm}
\usepackage{array}
\usepackage{textcomp}
\usepackage{stfloats}
\usepackage{url}
\usepackage{verbatim}
\usepackage{graphicx}
\usepackage[utf8]{inputenc}
\usepackage[T1]{fontenc}
\usepackage{graphicx}
\usepackage{caption}
\usepackage{subcaption}
\usepackage{makecell}

\usepackage{graphics} 
\usepackage{epsfig} 
\usepackage{xcolor}
\usepackage[normalem]{ulem}
\usepackage{booktabs}
\usepackage{multicol}
\usepackage{multirow}
\usepackage{cancel}
\usepackage{cite}
\usepackage{cleveref}
\usepackage{comment}
\usepackage{siunitx}
\let\labelindent\relax
\usepackage[inline]{enumitem}

\input{symbol_definitions}

\begin{document}

\title{Evaluation of Human-Robot Interfaces\\based on 2D/3D Visual and Haptic Feedback\\for Aerial Manipulation}

\author{Julien Mellet$^{1}$, Mike Allenspach$^{2}$, Eugenio Cuniato$^{2}$, Claudio Pacchierotti$^{3}$, Roland Siegwart$^{2}$, Marco Tognon$^{3}$
\thanks{$^{1}$PRISMA Lab, Department of Electrical Engineering and Information
Technology, University of Naples Federico II, Via Claudio 21, 80125, Naples, Italy.
Corresponding author's e-mail: {\tt\small julien.mellet@unina.it}.}%
\thanks{$^{2}$Autonomous Systems Lab (ASL), ETH Zurich.}%
\thanks{$^{3}$Univ Rennes, CNRS, Inria, IRISA - UMR 6074, F-35000 Rennes, France}%
\thanks{This work involved human subjects in its research. This study was conducted in accordance with the ethical principles outlined by the ETH Zurich Ethics Commitee. The research protocol and procedures were reviewed and approved as proposal No. EK-2023-N-81. All participants provided informed consent before participating in the study.}
\thanks{The research leading to these results has been supported by the AERO-TRAIN project, European Union's Horizon 2020 research and innovation program under the Marie Skłodowska-Curie grant agreement No 953454. The authors are solely responsible for its content.}}




\maketitle

\begin{abstract}
Most telemanipulation systems for aerial robots provide the operator with only 2D screen visual information. The lack of richer information about the robot's status and environment can limit human awareness and, in turn, task performance. While the pilot's experience can often compensate for this reduced flow of information, providing richer feedback is expected to reduce the cognitive workload and offer a more intuitive experience overall. This work aims to understand the significance of providing additional pieces of information during aerial telemanipulation, namely (i) 3D immersive visual feedback about the robot's surroundings through mixed reality (MR) and (ii) 3D haptic feedback about the robot interaction with the environment.
To do so, we developed a human-robot interface able to provide this information. First, we demonstrate its potential in a real-world manipulation task requiring sub-centimeter-level accuracy. Then, we evaluate the individual effect of MR vision and haptic feedback on both dexterity and workload through a human subjects study involving a virtual block transportation task. Results show that both 3D MR vision and haptic feedback improve the operator's dexterity in the considered teleoperated aerial interaction tasks.
Nevertheless, pilot experience remains the most significant factor.
\end{abstract}

\def\abstractname{Note to Practitioners}
\begin{abstract}
This study addresses the challenge of enhancing operator performance in telemanipulation applications with aerial robots. Traditional reliance on 2D screen displays can impair human cognitive awareness and task efficiency for inspection, maintenance, and complex assembly operations. Our proposed solution incorporates 3D mixed reality (MR) goggles and a haptic joystick to enrich the operator's perception. This human-robot interface proves being particularly advantageous for tasks requiring sub-centimeter-level accuracy in manipulation. The human subjects study demonstrates the practical benefits of using MR and/or haptics independently, showing that pilot dexterity is enhanced. However, pilot experience remains the most significant factor. Notably, transfer learning from the augmented setup to the standard one has been observed. Future research will investigate the learning rate associated with using such systems compared to current training techniques.
\end{abstract}


\section{Introduction}
\IEEEPARstart{W}{hile} small-scale aerial robots have traditionally been employed for tasks like surveillance and aerial imaging, there is a growing demand for these platforms to engage in physical interactions~\cite{active-interaction, past-future-am, redundant-am}. 
This is particularly pertinent in situations where employing human workers might be risky or impractical, such as inspecting remote or hazardous environments, performing maintenance on tall structures, or handling dangerous substances~\cite{am-lit, survey-am, mr-haptic-nuclear-waste}. 
Accordingly, new \acp{OMAV} have been presented~\cite{hamandi2021design}, which are able to generate thrust in all directions and independently control both position and orientation.
Their intrinsic capacity to generate thrust in all directions allows them to hover in arbitrary orientations, with independent control of both position and orientation. These vehicles are thus capable of precise motion and interaction force control, while effectively rejecting disturbances \cite{odar, brunner2022planning}.

\begin{figure}[tp]
  \centering
  \includegraphics[width=\linewidth]{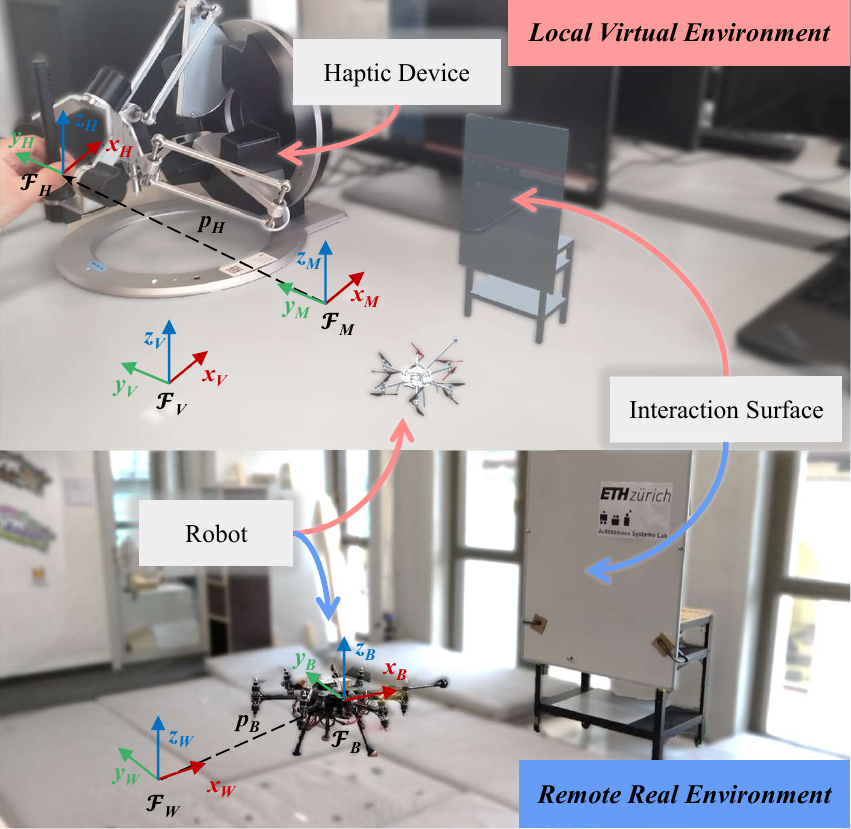}
  \caption{On the top: Local teleoperation environment with haptic joystick and hologram of the robot in its environment. On the bottom: Remote environment with the real robot.}
  \label{fig:eye-catcher}
  \vspace{-0.5cm}
\end{figure}

Although progress in the autonomous control of these vehicles is significant~\cite{6d-interaction, active-interaction, ndt}, 
for safety concerns, current regulations require a human operator to supervise any mission~\cite{aeroarms, autonomous-limitations}.
Rich sensory feedback is crucial for robotic teleoperation as it enables the human operator to perceive and interact with the remote environment effectively, thereby improving situational awareness and the capacity to perform complex tasks. Achieving a convincing sense of telepresence, creating the illusion of being physically present at a remote site, depends on advanced technology. 
Dedicated teleoperation systems facilitate the transfer of human expertise, decision-making, and cognitive abilities to remote environments.

A carefully designed teleoperation framework must ensure human situational awareness, by providing visual and/or haptic feedback, and an intuitive way to send reference commands to the remote robot.
The enhancement of perceptual awareness of the surroundings contributes to the concept of \textit{bilateral teleoperation}~\cite{bilateral}. A typical installation is shown in Fig.~\ref{fig:eye-catcher}.

While state-of-the-art bilateral teleoperation proved the potential of haptic and vision devices for free-flight applications, no prior research has investigated the influence of these technologies on operator performance in the aerial manipulation context~\cite{virtual-inertial-2-sam, mr-mav-haptic}.

\subsection{Related Work}
\label{related_work}


\begin{figure*}[thpb]
  \centering
  \includegraphics[width=\linewidth]{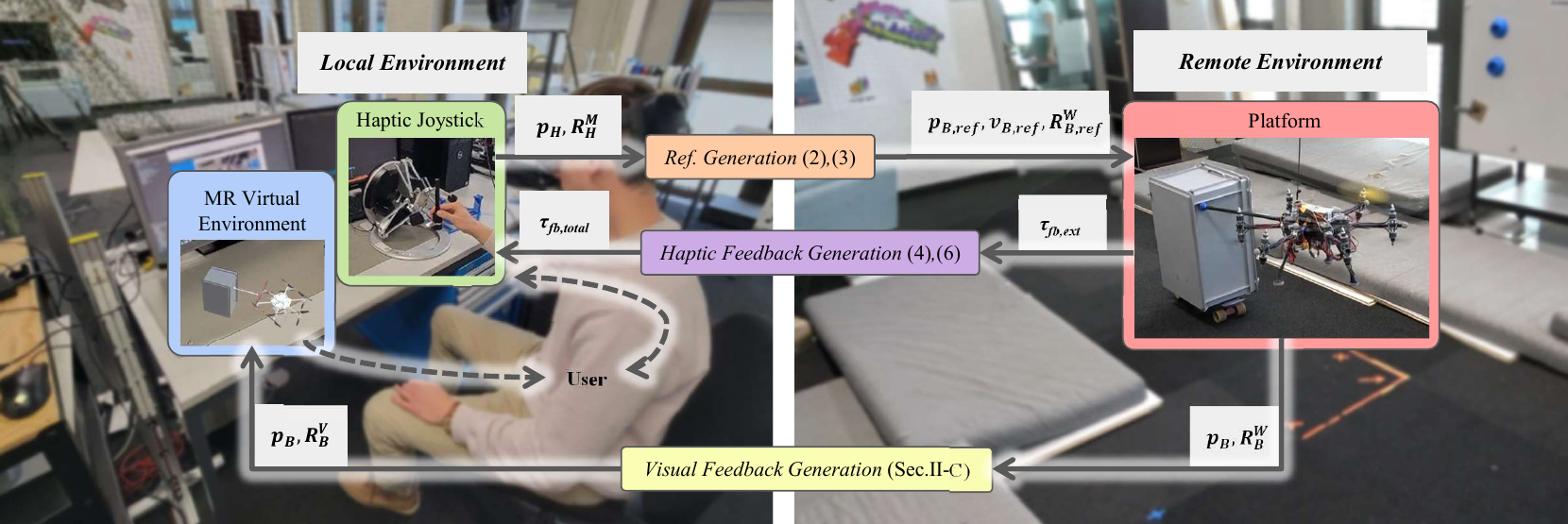}
  \caption{Detailed interactions between the different components of the teleoperation framework.}
  \label{fig:system}
  \vspace{-0.5cm}
\end{figure*}

Most state-of-the-art aerial teleoperation relies on 2D visual displays, providing limited information regarding the status of the robot and the surrounding environment. To enhance situational awareness, researchers proposed the use of immersive \ac{VR}~\cite{tactile-feedback ,sam-vio-vr, drone-pick} and \ac{MR} technology. The latter is especially promising as it enables to combine virtual and real content in a seamless way~\cite{real-time-mr, assisted-mr, mr-haptic-nuclear-waste}.
Differently from \ac{VR}, in which the user is fully immersed in a virtual environment, \ac{MR} enriches the real world with virtual content, combining real environment awareness with rich feedback information~\cite{vr-embodiement}.
%
%
However, \ac{MR} technology has mainly been used in scenarios where operators maintain direct sight on the remote environment~\cite{mr-surrogates, mr-traj-omav}.
Unlike setups with 2D displays for remote observation, \ac{MR} integration in the local context lacks exploration. 


Human perception, inherently multimodal, involves vision and proprioception, which is known to enhance performance~\cite{feel-vs-vision}. Thus, haptic feedback represents another relevant source of feedback shown to be effective in providing information in various teleoperation scenarios, including aerial robots~\cite{haptics-survey}.
Haptics technology provides touch-related information about the interaction of the robotic system with the remote environment or its virtual reconstruction~\cite{sam-vio-vr}. It was mostly used in contact-free flight \cite{free-flight-1, free-flight-2} or simulated physical interaction \cite{early-bilateral}. In aerial manipulation tasks, interaction forces play a crucial role in enhancing user control and manipulation capabilities~\cite{mike-6dof, rotation-ctl}.
Haptic feedback would allow users teleoperating an aerial manipulator to better perceive object properties, exert appropriate forces during manipulation, and enhance precision.
To obtain force feedback, various haptic devices, such as wearables \cite{tactile-feedback, wearable-haptics}, robot arms \cite{mike-6dof}, or force-feedback joysticks \cite{rotation-ctl} have been used.

Given the expected benefits of providing \ac{MR} and haptic feedback in the teleoperation of aerial manipulators, it is relevant to explore and evaluate which type of feedback is best suited for conveying what type of information. Indeed, most works on the topic either do not focus on evaluating \ac{HRI} performance~\cite{virtual-inertial-2-sam}, or only consider free-flight application~\cite{free-flight-1, free-flight-2, mr-mav-haptic}.

In summary, evaluating the effect of haptic and/or MR feedback on operator performance during aerial manipulation tasks remains an open research question. The importance or interconnection of operator multimodal sensing for aerial manipulation remains unexplored.

\subsection{Contributions}
This work aims at investigating the impact of sensory feedback, notably visual and haptic feedback, on user's experience and task performance during teleoperated aerial physical interaction. The main contributions of our work can be summarized as:
\begin{enumerate*}[label=\textit{\roman*)}]
    \item the design of a sensory interface coupling \ac{MR} and haptic feedback for 6-\ac{DoF} teleoperation,
    \item qualitative evaluation of the proposed framework in real-world benchmark test with sub-centimeter-level-accuracy interaction tasks, and 
    \item quantitative evaluation in a user study within a simulated-world interaction task.
\end{enumerate*}    

The rest of the paper is organized as it follows. Section~\ref{sec:system} presents the proposed system and each of its components. Then, Section~\ref{sec:experiment} shows experiments where the entire pipeline runs in a real-world application. Description of the quantitative evaluation scenario is done in Section~\ref{sec:user-study} with the user study. Finally, the results are discussed in Section~\ref{sec:discussion}.

\section{SYSTEM}
\label{sec:system}

The bilateral teleoperation system under study comprises four main elements:
\begin{enumerate*}
    \item \textit{human operator},
    \item \textit{haptic device}, 
    \item \textit{\ac{MR} device}, and
    \item \textit{aerial robot}.
\end{enumerate*}
The human operator, the \ac{MR} device and the haptic device are in the local environment, while 
the robot is located in the remote environment and communicates through a link with the local environment. We assume to know the environment. 
The full system is shown in Fig.~\ref{fig:system} and its components are discussed in detail in the following sections.



\subsection{Aerial Robot}
\label{omav_control}

The aerial robot utilized in this work is the tiltrotor \ac{OMAV} proposed in~\cite{active-interaction}.


The remote configuration is defined with the inertial world frame $\frameW=\{\originW,\xW,\yW,\zW\}$ with origin $\originW$ and unit axis $\{\xW,\yW,\zW\}$ located at an arbitrary fixed position such that $\zW$ is opposite to gravity. The state of the aerial robot is described by the body frame $\frameB=\{\originB,\xB,\yB,\zB\}$ with origin $\originB$ coinciding with robot's \ac{COM} and $\xB$ pointing toward the end-effector. To define the system's state, let $\pos \in \nR{3}$ express the position of $\originB$ with respect to $\frameW$. The attitude of $\frameB$ with respect to $\frameW$ is described by the rotation matrix $\rotMatWB \in \SO{3}$.
Finally, we denote the body linear and angular velocities with respect to $\frameW$ as $\vel,\;  \angVel \in \nR{3}$, respectively. For compliance at interaction, the robot is controlled by an impedance controller~\cite{active-interaction} together with a momentum-based external wrench estimator, which makes the closed-loop dynamics as
\begin{equation}
    \label{haptic_model}
    \Vec{M}_{v} \begin{bmatrix} \accB \\ \angVelB \end{bmatrix} + \Vec{D}_{v} \begin{bmatrix} \Vec{e}_v \\ \Vec{e}_{\omega} \end{bmatrix} + \Vec{K}_v \begin{bmatrix} \Vec{e}_p \\ \Vec{e}_{R} \end{bmatrix} = \wrenchExt,
\end{equation}
with $\mathbf{M}_{v}$, $\mathbf{D}_{v}$, $\mathbf{K}_{v} \in \nR{6\times6}$ the virtual inertia, the damping and the stiffness, respectively, which are tuning parameters. External wrench acting on the platform is described with $\wrenchExt \in \nR{6}$.
Finally, we consider the position $\Vec{e}_p$, orientation $\Vec{e}_R$, velocity $\Vec{e}_v$ and angular rate $\Vec{e}_{\omega}$ errors as, 
\begin{subequations}
 	\label{position_errors}
 	\begin{IEEEeqnarray} {ll}
 		\Vec{e}_p &= \rotMatWB {\transpose} \left( \pos - \posRef \right),\\
         \Vec{e}_R &= \frac{1}{2}\left( \rotMatWBRef{\transpose} \rotMatWB - \rotMatWB {\transpose} \rotMatWBRef \right)^\vee,\\
         \Vec{e}_v &= \rotMatWB {\transpose} \left( \vel - \velRef \right),\\
         \Vec{e}_{\omega} &= \Vec{\omega}_B - \rotMatWB {\transpose}  \rotMatWBRef \Vec{\omega}_{B,ref},
 	\end{IEEEeqnarray}
\end{subequations}
with $\left(\cdot\right)^\vee$ the Vee operator to extract a vector from a skew-symmetric matrix.
The reference position and orientation, 
$\posRef$ and $\rotMatWBRef$, respectively, are computed as in Sec.~\ref{sec:haptics}. 
This allows us to steer the aerial robot and safely interact with the remote environment. 



\subsection{Haptic System and Control}
\label{sec:haptics}
Through the haptic interface, the operator controls the vehicle's velocity. The inertial frame $\frameM=\{\originM,\xM,\yM,\zM\}$ with origin $\originM$ corresponds to the idle position and orientation of the joystick's handle (see Fig.~\ref{fig:eye-catcher}). Its current pose is described by the frame $\frameH=\{\originH,\xH,\yH,\zH\}$ with origin $\originH$ fixed with respect to the handle. We denote position and velocity of the haptic device handle with $\posH \in \nR{3}$ and $\velH  \in \nR{3}$, both expressed with respect to $\frameM$. 
Attitude and angular rate of $\frameH$ with respect to $\frameM$ are defined as $\rotMatMH \in \SO{3}$ and $\angVel \in \nR{3}$, respectively, the latter expressed in $\frameH$.


\subsubsection{Reference Generation}

Similarly to~\cite{mike-6dof, wearable-haptics}, to create an intuitive correspondence between the limited input workspace and the potentially boundless robot workspace, we generate velocity commands through the haptic interface, as it represents a widely accepted approach for teleoperating aerial vehicles~\cite{finite-to-infinite}. Translational and rotational references are then computed such that,
\begin{subequations}
    \label{translation_ref}
    \begin{IEEEeqnarray} {ll}
        \velRef &= \velMax \posH,\\
        \posRef &= \int_{0}^{t} \velRef(b) \,db,\\
        \Vec{\omega}_{B,ref}^B &= \frac{\omega_{max}}{2} (\rotMatMH - {\rotMatMH}^{\transpose})^{\vee},\\
        \rotMatWBRef &= \int_{0}^{t} \rotMatWBRef(b) \left[ \Vec{\omega}_{B,ref}(b) \right]_{\times} \,db,
	\end{IEEEeqnarray}
\end{subequations}
where $(\cdot)_{\times} : \nR{3} \mapsto \SO{3}$ is the skew-symmetric operator, $\velMax$ and $\omega_{max}$ are respectively velocity and angular rate gains set by the operator preference.

\subsubsection{Haptic Feedback Generation}
As proposed in~\cite{mike-6dof}, we define the total feedback wrench as,
\begin{equation}
    \label{total-wrench}
    \wrenchTotal = \wrenchRec + \wrenchFbExt,
\end{equation}
where $\wrenchRec \in \nR{6}$ is the recentering wrench and $\wrenchFbExt \in \nR{6}$ is the interaction wrench. In details,

\begin{equation}
    \label{recentering-wrench}
    \wrenchRec = -\mathbf{K}_{rec} \begin{bmatrix} \posH \\ \frac{1}{2} (\rotMatMH - {\rotMatMH}^{\transpose})^{\vee} \end{bmatrix},
\end{equation}
with $\mathbf{K}_{rec} = diag(\mathbf{K}_{rec,t} , \mathbf{K}_{rec,r}) \in \nR{6 \times 6}$ a tuning parameter. When the operator release the joystick, thanks to the recentering action, it returns to the idle position, corresponding to a zero robot velocity. Moreover, the wrench feedback at interaction is defined as,
\begin{equation}
    \label{recentering-wrench}
    \wrenchFbExt = \mathbf{K}_{ext} \wrenchEst,
\end{equation}
with $\mathbf{K}_{ext} \in \nR{6}$ a tuning parameter to adjust perceived effort to user preferences. 
Since the described haptic feedback only provides information at contact, the visual feedback is added to improve spatial and depth perception.

\subsection{Mixed Reality Vision} \label{sec:MR}
The \ac{MR} headset creates holograms with respect to a virtual frame $\frameV=\{\originV,\xV,\yV,\zV\}$ that can be set arbitrarily with origin $\originV$.

\begin{figure} 
    \centering
    \subfloat[Remote push\label{fig:experiment-a}]{%
        \includegraphics[width=0.49\linewidth]{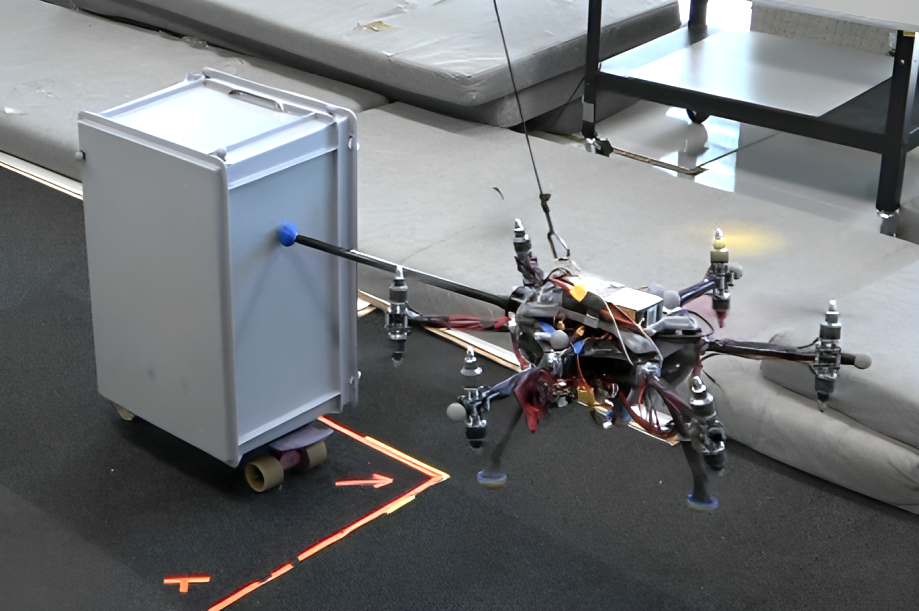}}
    \hfill
    \subfloat[Remote insertion\label{fig:experiment-b}]{%
        \includegraphics[width=0.49\linewidth]{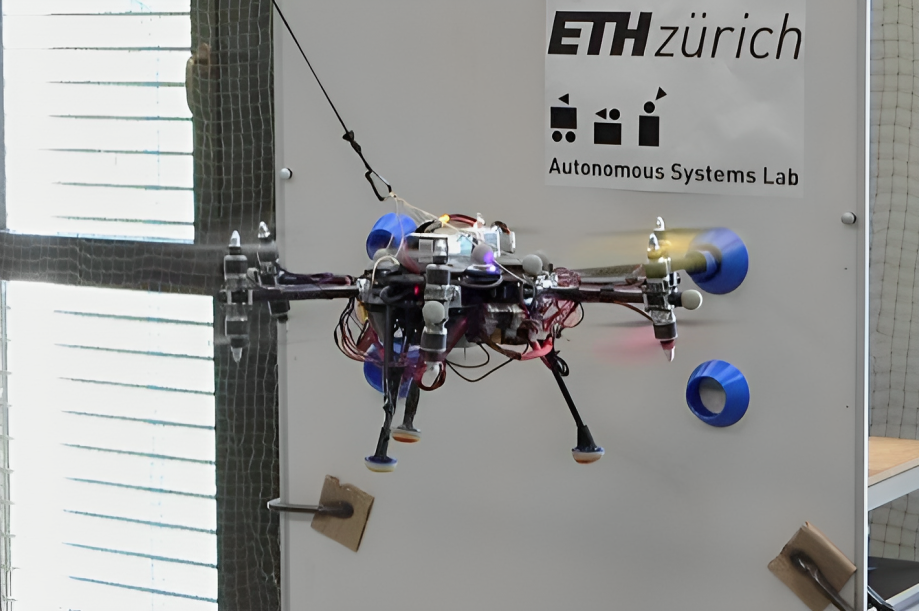}}
    \\
    \subfloat[Virtual push\label{fig:experiment-c}]{%
        \includegraphics[width=0.49\linewidth]{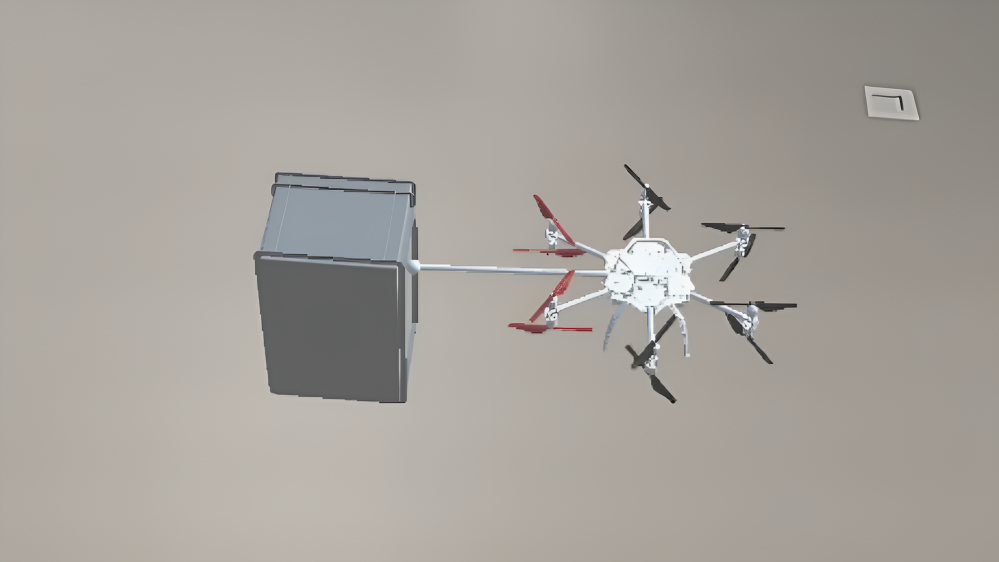}}
    \hfill
    \subfloat[Virtual insertion\label{fig:experiment-d}]{%
        \includegraphics[width=0.49\linewidth]{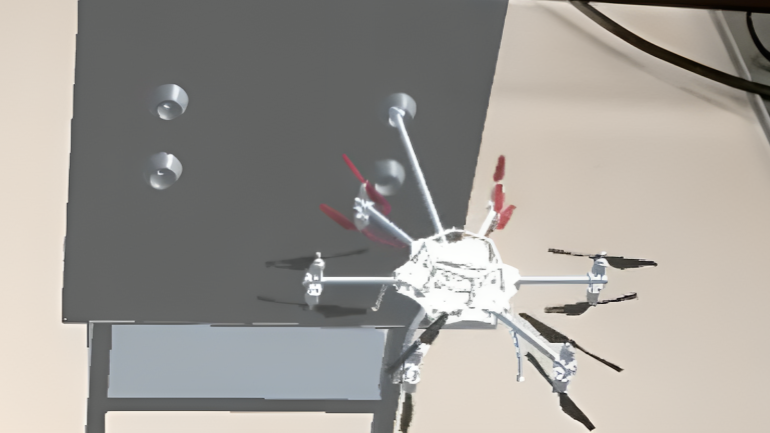}}

  \caption{Experimental setup for force exertion~\ref{fig:experiment-a} and insertion task~\ref{fig:experiment-b}, with corresponding virtual \ac{MR} holograms~\ref{fig:experiment-c} and~\ref{fig:experiment-d}.}
  \label{fig:experiments}
  \vspace{-0.5cm}
\end{figure}

The operator sees the robot in its environment through the \ac{MR} goggles. The virtual frame $\frameV$ is constant with respect to the inertial local frame $\frameM$, thanks to the localization features implemented on the goggles. Holographic visualizations are designed based on the task at hand. Essential information includes the position and orientation of the robot and interaction objects. Examples for peg-in-hole, pushing, and pick-and-place can be found in Figs.~\ref{fig:experiment-c},~\ref{fig:experiment-d} and~\ref{fig:setup-b}, respectively.





\section{SYSTEM DEMONSTRATION}\label{sec:experiment}

\begin{figure*}[thpb]
  \centering
  \includegraphics[width=\linewidth]{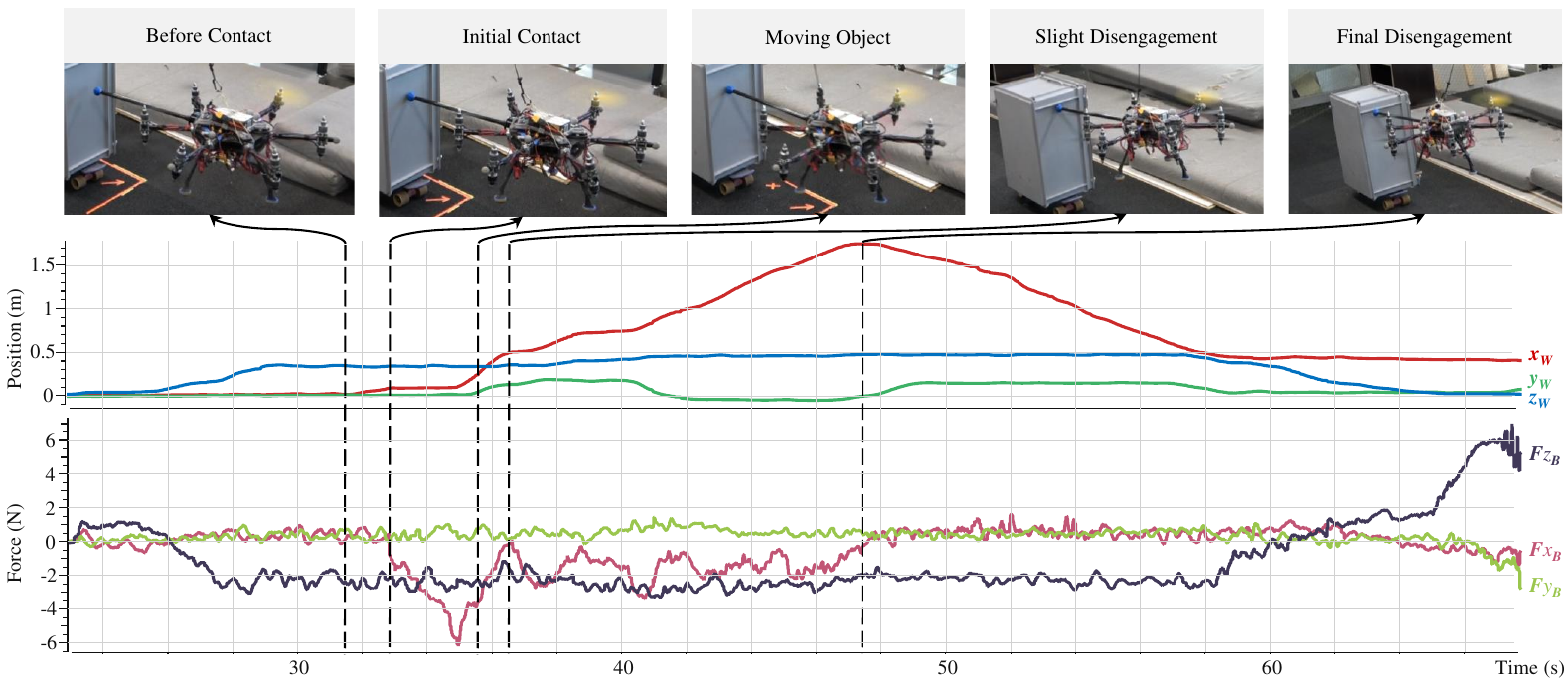}
  \caption{Experiment 1 - Measured position in $\frameW$ and estimated forces in $\frameW$ while pushing the wheeled object. From left to right the highlighted moments are: before the interaction, the first physical contact, a slight discontinuity of contact, and the final contact disengagement.}
  \label{fig:plot-push-a-box}
\end{figure*}
\begin{figure*}[thpb]
  \centering
  \includegraphics[width=\linewidth]{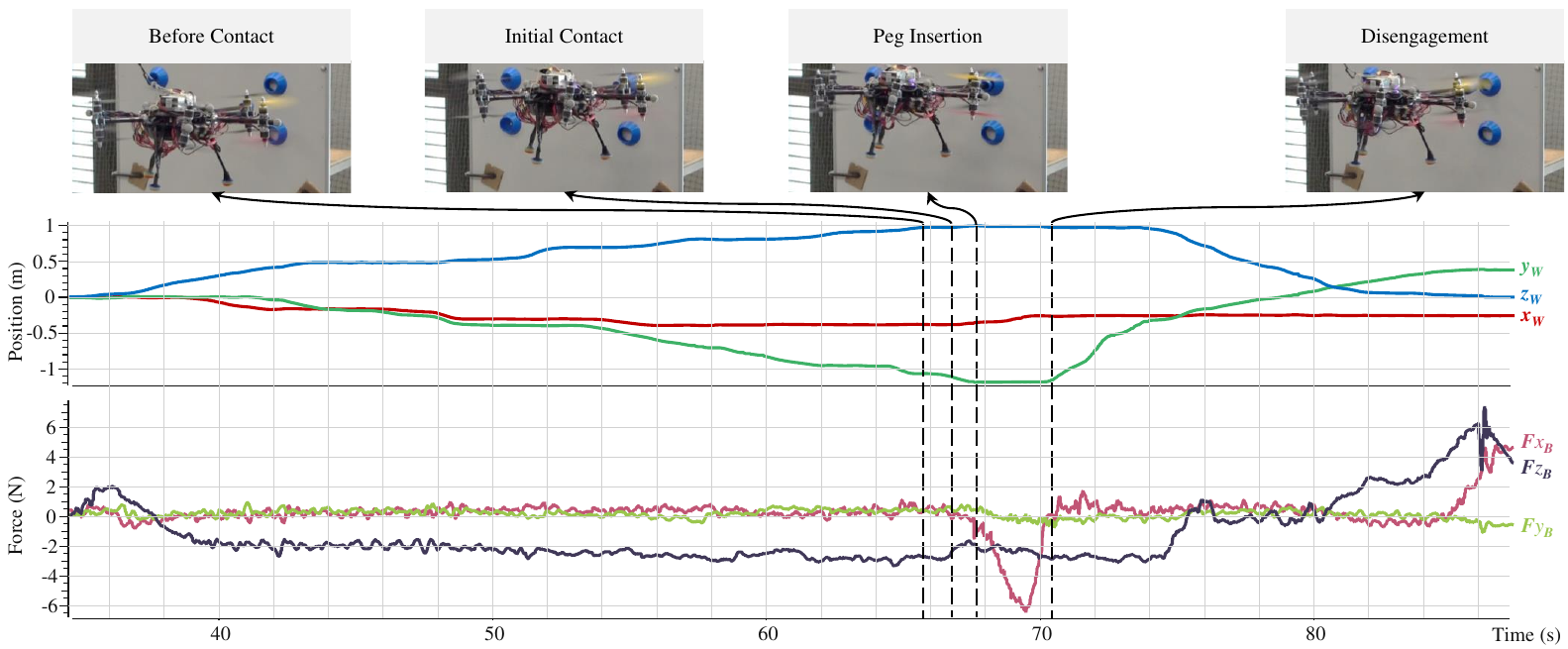}
  \caption{Experiment 2 - Measured position in the world frame $\frameW$ and estimated forces in the body frame $\frameB$ during the peg-in-hole task. From left to right the highlighted moments are: before the interaction, the shocks before the insertion, the reach of the hole's end, and the exit from the hole.}
  \label{fig:plot-peg-in-hole}
  \vspace{-0.2cm}
\end{figure*}


First, we show the viability of using the proposed approach in a teleoperation benchmark for the aerial robot, teleoperated in physical interaction with a remote environment. 
The goal is to show the feasibility of controlling the aerial robot in a real physical setup, with data streamlines through a network, while the operator only gets information through \ac{MR} along with the haptic joystick.

\subsection{System Setup}\label{sec:experiment-setup}
We propose two scenarios with physical contact, in which force feedback information for the operator is essential to ensure safe interaction:
\begin{enumerate*}
    \item pushing a wheeled object (in paragraph~\ref{sec:experiment-force-exertion}), and 
    \item inserting a peg into a hole (in paragraph~\ref{sec:experiment-insertion}).
\end{enumerate*}
The potential of haptic feedback for exerting force on movable objects, along with the potential of 3D visual feedback for high-accuracy insertion tasks, are both significant in terms of achieving task objectives. 
The proposed scenarios have been chosen as they require rich visual and force feedback information to be well executed. 
They also show high flexibility of the framework for different tasks.

Preliminary flight tests have been used for gain tuning. For safety reasons, we limit the maximum velocity of the robot to $ \velMax = $ 15\,\si{c\meter\per\second}. Moreover, the force feedback was scaled to one third of the measured force, to account for the maximum force output of the device ($12$~N) and avoid any saturation of the force feedback. 
These adjustments allow the operator to perceive external forces while reducing the physical effort he exerts.
%
The robot is teleoperated by an experienced operator who has no direct view of the remote environment.

For all these experiments, the interaction surface and the aerial robot poses are tracked with a motion capture system. The real environment is known a priori and replicated in the virtual space through the Microsoft HoloLens 2 goggles~\cite{hl2}. 

\paragraph{Experiment 1 - Pushing a Wheeled Object}\label{sec:experiment-force-exertion}
This task is relevant because it combines flight accuracy, force exertion regulation, and 3D localization awareness. In particular, visual feedback is required to make the contact with the box in the center, and haptic feedback enables to apply the right amount of force for keeping the box in motion without damaging it.
In this scenario, a box is mounted on a wheeled platform and the aerial robot needs to apply varying forces to overcome static friction, to then keep contact. Figures~\ref{fig:experiment-a} and~\ref{fig:experiment-c} present the aerial platform pushing the box.

\paragraph{Experiment 2 - Inserting a Peg Into a Hole}\label{sec:experiment-insertion}
The second task consists of fitting a cylindrical peg into a corresponding hole attached on a vertical interaction surface (see Fig.~\ref{fig:experiment-b}). This task is relevant because it combines flight accuracy and fine manipulation. Visual feedback is important to provide a close-up view of the hole, while haptic feedback helps during insertion (e.g.,~\cite{cutaneous-stability}). The tip radius measures $20$~mm while the hole is $25$~mm. The task thus requires a $5$~mm accuracy from the operator to be completed.

\subsection{Results and Analysis}\label{sec:results}
The experiments demonstrate the system's capability to provide intuitive and precise control in complex tasks. Below, we present the detailed outcomes and analyses of these experiments.

\paragraph{Experiment 1}
The outcome of the first experiment, which involved pushing the object, demonstrates that the operator managed to regulate the robot's force using the haptic device, as shown in Fig.~\ref{fig:plot-push-a-box}. The aerial robot applied a force of roughly $5$~N to overcome the static friction and pushed the box approximately $1.7$~meters. During the interaction, the system experienced a maximum estimated force of up to $5.2$~N while maintaining stable contact with an average force of $1.0$~N to overcome the kinetic friction.

\paragraph{Experiment 2}
The insertion task was successfully accomplished by fitting the peg into the narrow hole. The operator required less than $5$~mm lateral accuracy in operating the robot. With the 3D virtual representation, the user was able to ensure the alignment of the robot with the target by positioning his head around the scene. At the moment of insertion (see Fig.~\ref{fig:plot-peg-in-hole}), the force along the manipulator increased up to $5.5$~N. The task was completed rapidly, taking a total of $2.5$~s inside the hole, avoiding any major interactions between the rigid bodies. The lateral force in the normal plane of the hole was regulated by the operator and anticipated such that no peak was noticeable in the force measurement. This allowed the system to achieve sub-centimetric accuracy in insertion tasks.\\

In the virtual space, one can notice surface collision of the end-effector mesh with the hole of the whiteboard in Fig.~\ref{fig:experiment-d}. This is due to some compliance of the board at contact. This was neglected, assuming a solid board. This visual artifact did not interfere with the operator's task, and found the overall system pleasant and natural to use. The haptic feedback has also been felt informative by the operator to understand task completion.

In the end, the experiments showed in Figs.~\ref{fig:experiments} proved the feasibility of using the system for \ac{APhI} tasks. Videos of these experiments are in the supplementary material. 
We demonstrated the capabilities afforded by combining haptic and visual feedback, specifically achieving precise control over force exertion with haptic feedback and accurate positioning and alignment with visual feedback.
The question raises if the operator would perform similarly on a physical task without haptics or \ac{MR}.

\section{USER STUDY}\label{sec:user-study}
After demonstrating the potential of combining rich haptic and visual feedback for teleoperating aerial manipulation tasks, we conduct a quantitative study to assess their impact on performance and perceived workload. As the role of sensory feedback also depends on the experience of the operator~\cite{expert-vs-beginner-operation}, we also investigate the influence of experience in this context. In conventional statistical design, optimal conditions are determined based on measured values. In contrast, the Taguchi design identifies optimal conditions by selecting parameters that exhibit the lowest variability. Here, the optimal refers to the most stable and consistent combination in term of task efficiency, robustness through repeatability (i.e., predictable outcomes even with variations in external factors) while reducing the workload.

\subsection{Design of Experiment}\label{sec:doe}
We considered three factors in our analysis, evaluating the role of the \textit{visual feedback}, \textit{haptic feedback}, and \textit{user's experience} in the teleoperation of an aerial manipulator.


The \textit{visual feedback} provides visual information to the operator about the status of the \ac{OMAV} (see Sec.~\ref{sec:MR} and Fig.~\ref{fig:user-study-setup}). We considered two different types of visual feedback:
    \begin{itemize}
        \item $SC$: a representation of the robot in a virtual environment using a standard 2-dimensional \ac{LCD} computer screen (see Fig.~\ref{fig:setup-a}),
        \item $MR$: a representation of the robot in a mixed-reality environment using the 3-dimensional \ac{MR} vision headset HoloLens 2 (see Fig.~\ref{fig:setup-b}).
    \end{itemize}
    
The \textit{haptic feedback} provides information to the operator about the interaction forces acting on the OMAV (see Sec.~\ref{sec:haptics}). We considered two haptic feedback conditions:
    \begin{itemize}
        \item $H$: providing force information at physical contact (i.e. $\wrenchFbExt$) through an Omega.6 grounded haptic interface,
        \item $\cancel{H}$: no haptic feedback, i.e., set $\wrenchFbExt=0$ in~\cref{total-wrench}.
    \end{itemize}
Finally, we also considered two levels of \textit{operator expertise}:
    \begin{itemize}
        \item $B$: beginner users, i.e., users who discovered the setup and task on the day of the experiment, 
        \item $E$: experienced users, i.e., users who followed a training before the experiment (see Sec.~\ref{sec:participants}).
    \end{itemize}
\textit{Visual feedback} and \textit{haptic feedback} are considered as within-subjects factors, while the \textit{operator expertise} is considered as a between-subjects factor.

\begin{table}[t]
    \small
    \centering
    \caption{Taguchi Design $L_4$ $(2^3)$}
    \label{tab:taguchi} 
    \begin{tabular}{r|ccc}
        \hline
        \multicolumn{1}{c|}{\multirow{2}{*}{\begin{tabular}[c]{@{}c@{}}Experimental\\ Conditions\end{tabular}}} & \multicolumn{3}{c}{Column (Factors)} \\
        \multicolumn{1}{c|}{} & \textbf{\begin{tabular}[c]{@{}c@{}}Display\\ Technology\end{tabular}} & \multicolumn{1}{l}{\textbf{Haptics}} & \textbf{\begin{tabular}[c]{@{}c@{}}Operator\\ Expertise\end{tabular}} \\ \hline
        1 & SC & \cancel{H} & B \\
        2 & SC & H & E \\
        3 & MR & \cancel{H} & E \\
        4 & MR & H & B \\ \hline
    \end{tabular}
\end{table}

\begin{table}[t]
    \small
    \centering
    \caption{Complementary Experiments}
    \label{tab:complementary} 
    \begin{tabular}{r|ccc}
        \hline
        \hspace{12mm} 5 & \hspace{5mm} SC \hspace{5mm} & \hspace{3mm} \cancel{H} \hspace{3mm} & \hspace{4mm} E \hspace{4mm} \\
        6 & MR & H & E \\ \hline
    \end{tabular}
\end{table}

As testing all the combinations would lead to an experiment that is very long (>1 hour), we minimize the number of tasks tested to prevent human fatigue and to maintain a full-rank task feature space by using the $L_4$ $(2^3)$ Taguchi table~\cite{taguchi, taguchi-1}. This is a version of fractional factorial design of experiment. Taguchi design lowers the number of tasks to 2 per participant and is presented in Tab.~\ref{tab:taguchi}.  The proposed table ensures that the beginner group performs on what we might call the \textit{standard teleoperation} interface (i.e., \cancel{H}-SC) and on the proposed augmented setup (i.e., H-MR). Experiments 2 and 3 are subsequent to the data defined above and complete the Taguchi table.


In addition to the conditions considered by the Taguchi design, we also tested a few additional combinations (see Tab.~\ref{tab:complementary}), as to better understand the relationship between the considered variables. \ac{ANOVA} is a popular collection of statistical models and estimation tools used to analyze differences among output means of a sample \cite{anova-measures}. By increasing the number of data, we aim to cover the factor levels more thoroughly, validate our initial findings, and reduce the risk of statistical errors. Ultimately, this will lead to more reliable insights for analysing our study. We apply these tools in Secs.~\ref{sec:data-collection} and~\ref{sec:discussion}. 



To reduce learning effect, we shuffled randomly and uniformly the experimentation order of the various experimental conditions per participants.

\subsection{Setup and Task Definition} \label{sec:task-definition}
\begin{figure}[t]
    \centering
    \subfloat[Visual feedback under the 2D-LCD condition\label{fig:setup-a}]{%
        \includegraphics[width=0.49\linewidth]{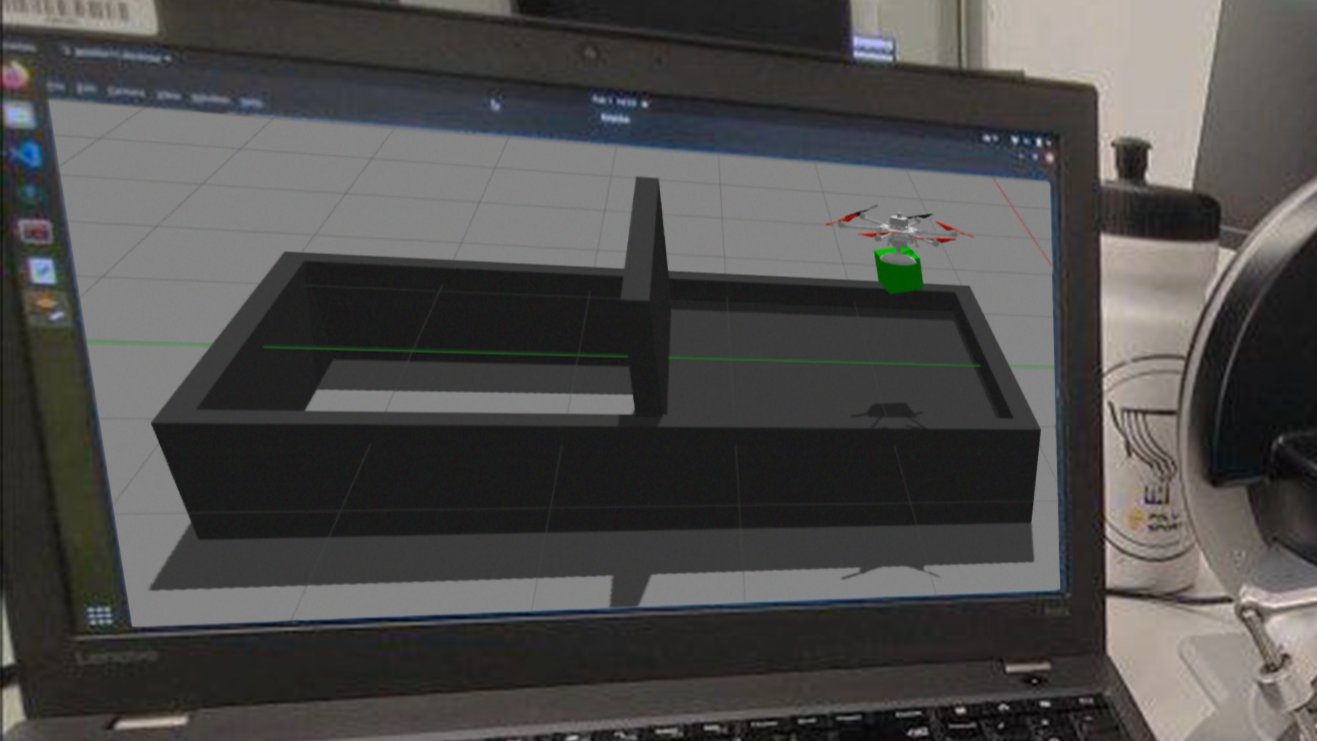}}
    \hfill
    \subfloat[Visual feedback under the 3D MR condition\label{fig:setup-b}]{%
        \includegraphics[width=0.49\linewidth]{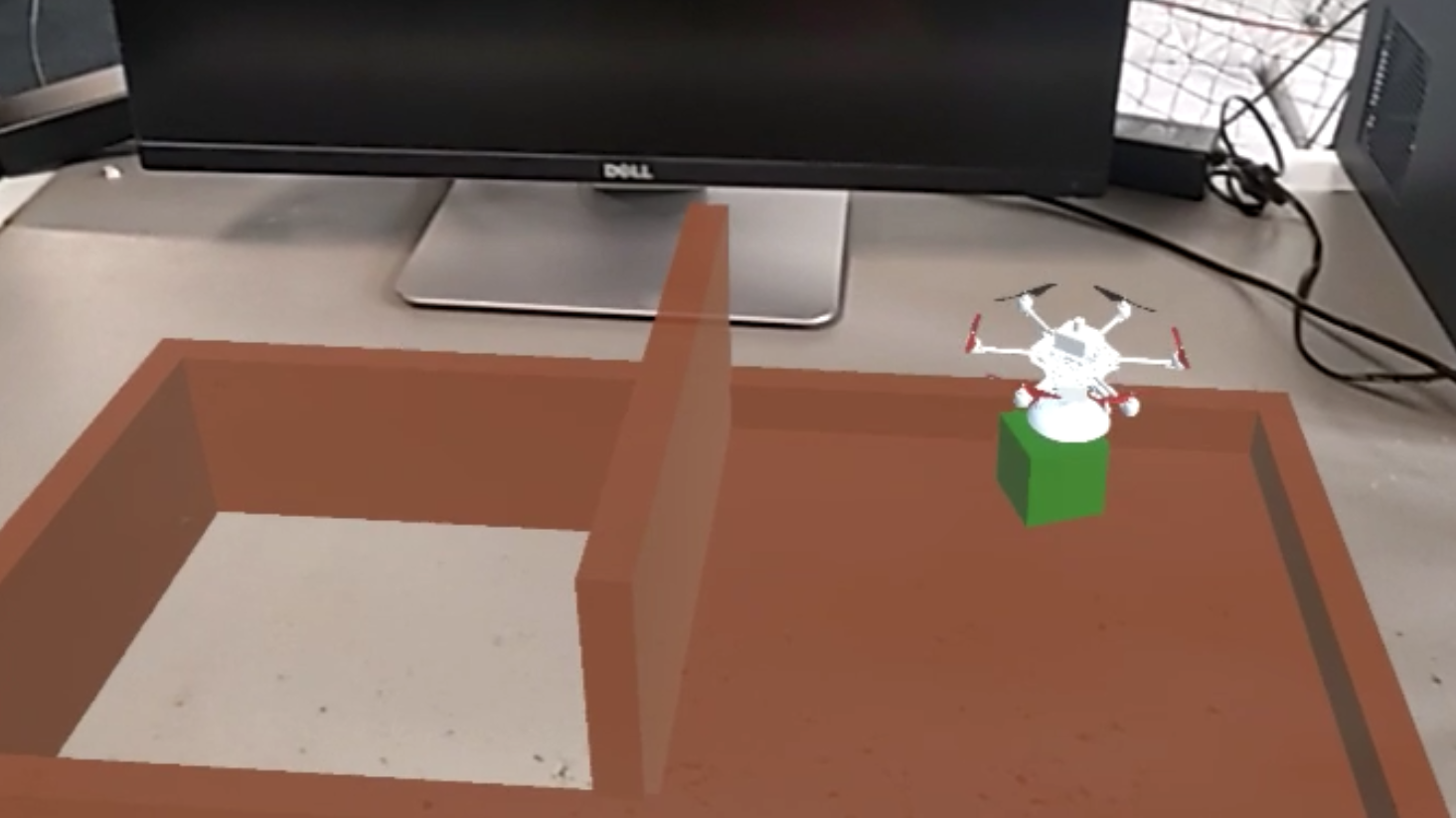}}
  \caption{User study display technologies showing block transported by the aerial robot in the \ac{ABBT} environment.}
  \label{fig:user-study-setup}
  \vspace{-0.5cm}
\end{figure}

Like in Sec.~\ref{sec:experiment}, the virtual environment appears to the operator thanks to the HoloLens 2 \ac{MR} headset or the \ac{LCD} screen display (see Fig.~\ref{fig:user-study-setup}), depending on the considered condition, while the remote robot is controlled through the Omega.6 haptic device in all cases.

Even if the proposed system enables full 6-\ac{DoF} control of the robot pose (see Sec.~\ref{sec:haptics}), for the purposes of the considered tasks, we decided to simplify the control of the robot keeping a horizontal attitude. The orientation of the vehicle is kept constant along $\xB$ and $\yB$ (see Fig.~\ref{fig:eye-catcher}). This means that the operator controls the translational velocities and the orientation around $\zB$ only.
Such an approach statistically improves the significance of the results, as rotational commands tend to decrease user performance~\cite{rotation-ctl}.
The operator sits in a posture similar to the one shown in Fig.~\ref{fig:system}. The remote environment is simulated in Gazebo with a physics engine~\cite{ode-engine}. The propellers' drag, lift coefficients and inertial unit characteristics are simulated through the internal plugin.

The \ac{BBT} is one of the most used and recommended test to evaluate human manual gross dexterity~\cite{bbt}. The participant is required to move blocks from one side of a partitioned box to the other within a specified time frame. It has been shown effective in real, virtual an mixes reality environments~\cite{BBT-VR2, BBT-VR4, ar-manip, vbbt}.
Because it has been successfully applied in~\cite{robot-bbt} we adapted the \ac{BBT} for \ac{APhI} in teleoperation, calling it \ac{ABBT}. The original clinical setup (Box and Blocks) has been scaled up of a factor 16, to account for the larger dimensions of the aerial manipulator. \ac{ABBT}, like \ac{BBT}, consists of pick and place blocks from one side of the box to another while avoiding the middle partition (see Fig.~\ref{fig:user-study-setup}). The virtual setup seen through the \ac{MR} goggles replicates the official test (see Fig.~\ref{fig:setup-b}), so that we keep the essence of the test using only a haptic device instead of the direct hand. Nevertheless, since the OMAV spatial displacement is slower than the hand displacement, we increased the task duration to $80$s.

\subsection{Participants}\label{sec:participants}
Twenty-eight participants (6 females, 22 males, aged 18-32) took part in the study, divided in two groups (Beginners and Experienced) with equal numbers, as discussed in Sec.~\ref{sec:doe}.

Experienced participants for this user study were selected based on their availability for a two-month training period, facilitating consistent and intensive training sessions.
Before training, we evaluated their prior knowledge: most were new to aerial robots, and all considered themselves fast learners. None had haptic device experience, a few had used \ac{VR} headsets, and none had prior \ac{MR} combined with haptics for \ac{APhI} exposure.

\begin{figure}[t]
    \centering
    \subfloat[Race track\label{fig:training-race}]{%
        \includegraphics[width=0.49\linewidth]{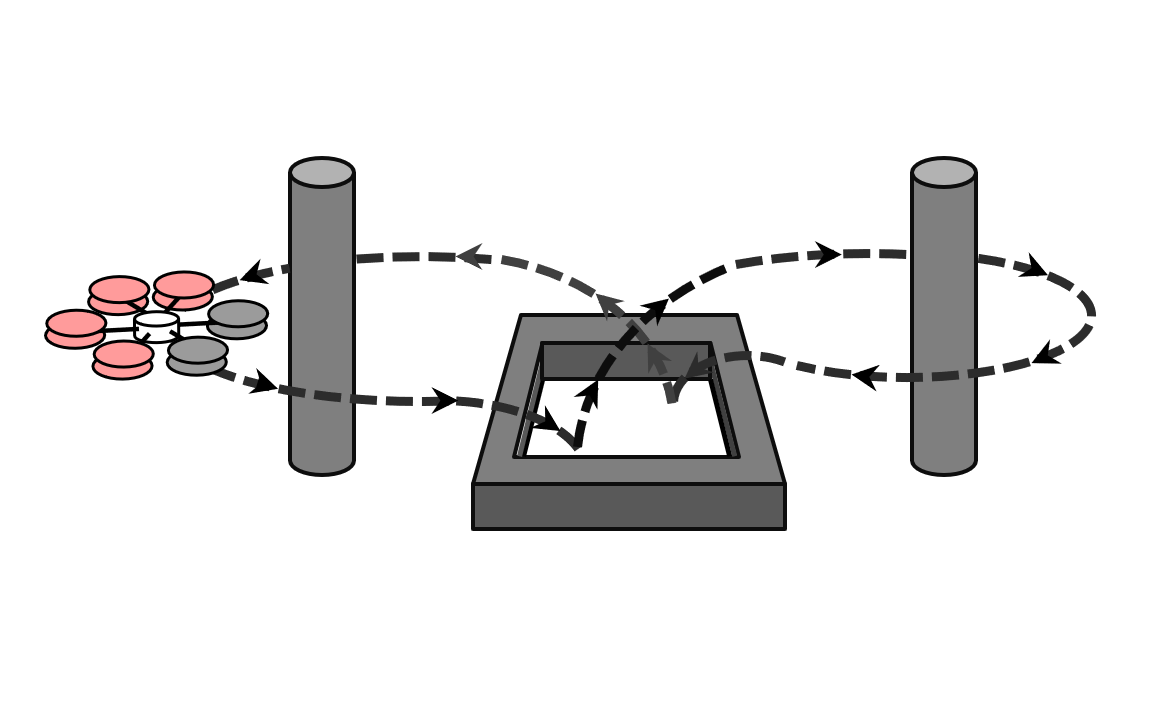}}
    \hfill
    \subfloat[Catch a ball\label{fig:training-catch}]{%
        \includegraphics[width=0.49\linewidth]{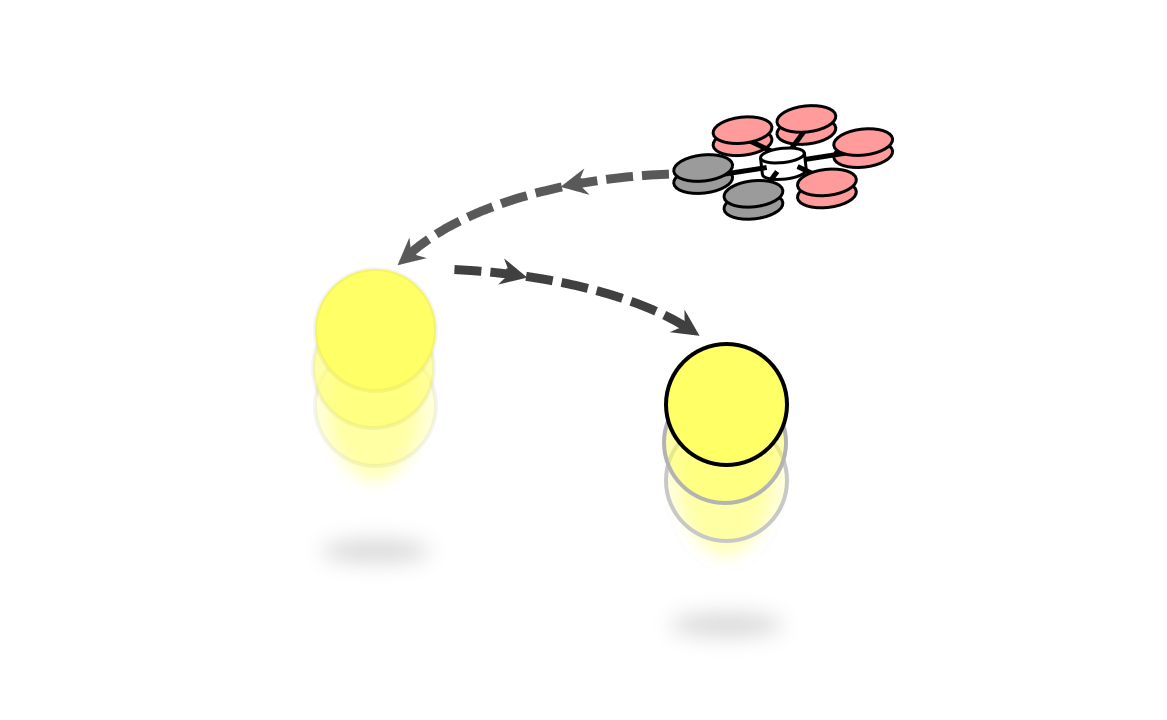}}
    \\
    \subfloat[Push a ball\label{fig:training-push}]{%
        \includegraphics[width=0.49\linewidth]{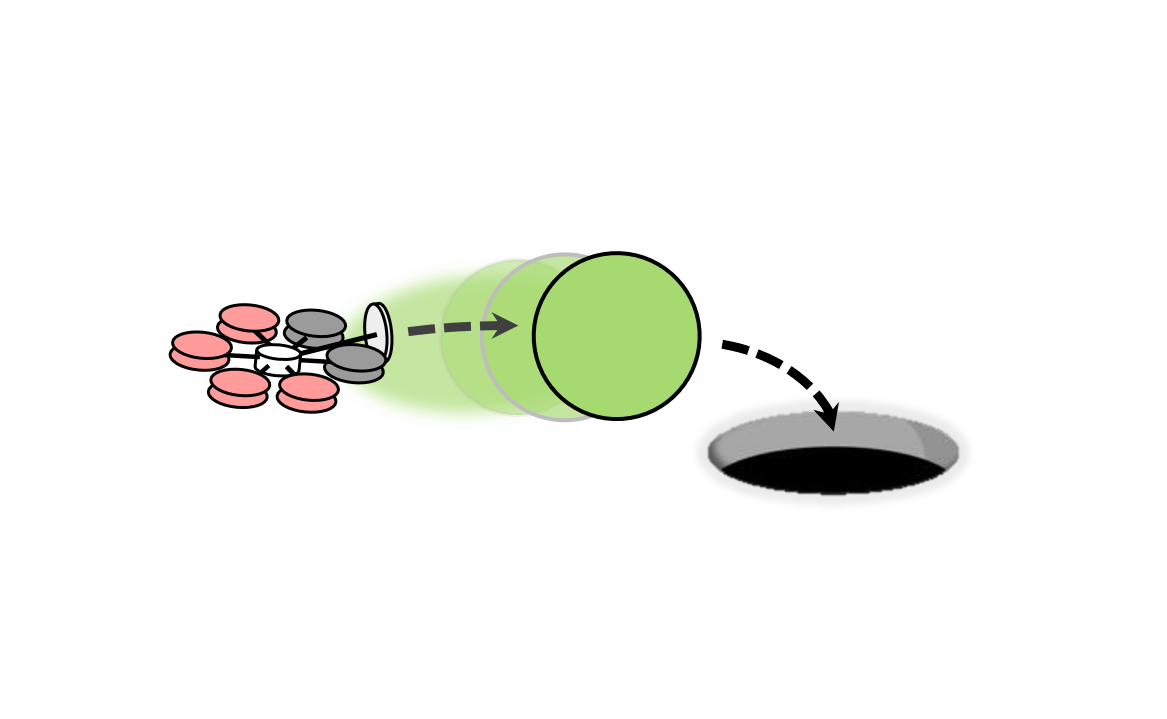}}
    \hfill
    \subfloat[ABBT\label{fig:training-abbt}]{%
        \includegraphics[width=0.49\linewidth]{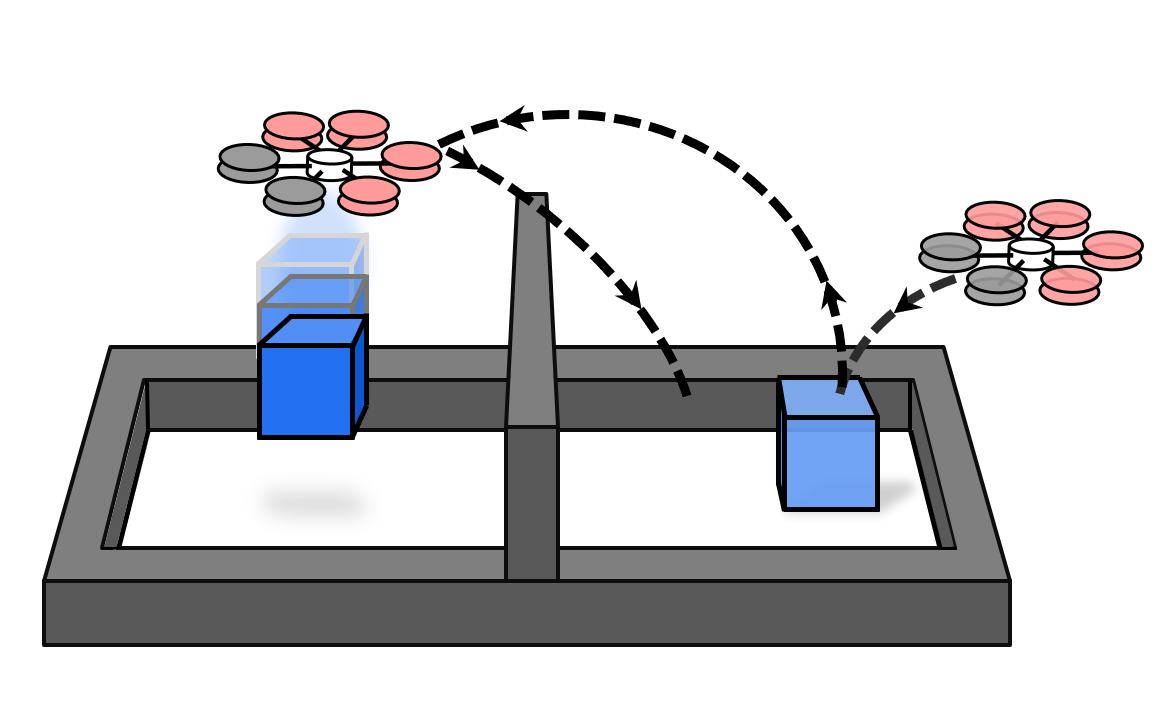}}
  \caption{Illustrative overview of the training activities (a)-(c) and the evaluation task (d).}
  \label{fig:training-set}
\end{figure}

\begin{figure}[t]
  \centering
  \includegraphics[width=\linewidth]{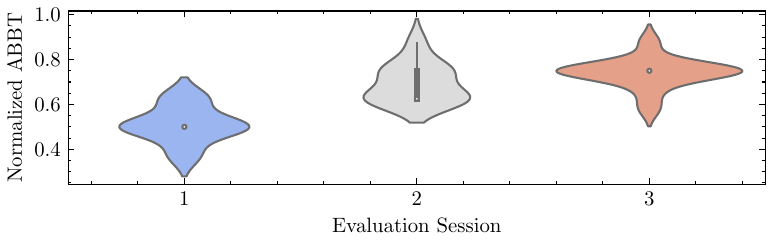}
  \caption{Evaluation of the experienced operator group during the training period. The violin plots illustrate the distribution of users performance with \ac{ABBT} evaluation session between training sessions.}
  \label{fig:training-evaluation}
\end{figure}

All training prior to the experimentation was conducted using our system, which combines haptics and \ac{MR}. We devised training activities to familiarize with the proposed multimodal framework. The activities included 
\begin{enumerate*}[label=\textit{\roman*)}]
    \item a ``\textit{Race Track}'' for improving aerial trajectories by slaloming between two poles and touching the ground between them (see Fig.~\ref{fig:training-race}),
    \item ``\textit{Catch a Ball}'' to enhance dexterity at catching a randomly jumping ball in mid-air (see Fig.~\ref{fig:training-catch}), and
    \item ``\textit{Push a Ball}'' for practicing force control, in an activity similar to golf, hitting a ball into a hole (see Fig.~\ref{fig:training-push}).
\end{enumerate*}
Videos of the activities are in the supplementary material. Over an 8-weeks period, participants underwent six 15-minutes sessions, with periodic evaluations using the \ac{ABBT} (see Figs.~\ref{fig:training-abbt} and~\ref{fig:training-evaluation}).

In the first evaluation session, the distribution is centered on the median at $0.5$, and has an interquantile range of $0.0$. By the second session, the distribution widens, showing a more important interquartile range of $0.125$, suggesting slight decrease in consistency but increase of the median to $0.625$. In the third session, the distribution narrows to an interquartile range of $0.0$ and median of $0.75$, indicating that users have continued to improve and stabilized their performance. The difference between session 1 and 3 indicates a significant variation in user performance. Overall, the data shows that user performance becomes more consistent and improves significantly over time with additional training. Notably, all participants exhibited improved telemanipulation skills, and were confident in the task by the end of the training period.

All participants reported no issues with visual or haptic perception. The experimenter provided detailed procedure explanations, taking about 5-minutes to ensure correct setup before the experiment for everyone. For the set of beginner users, subjects had another 5-minutes practice session to familiarize themselves with the telemanipulation system. No sensitive or personally identifiable data was collected, and protocols were approved by an ethical committee (No. EK-2023-N-81).

\subsection{Evaluation Metrics}
We considered three metrics:
\begin{itemize}
    \item The number of blocks transferred,  $\cal{N} \in \nN{}$, in the given $80$~\si{\second}. It evaluates the gross manual dexterity of the operator. Higher is better.
    \item The energy spent on block displacement $\cal{E} \in \nR{}$
        \begin{equation}
            {\cal E} = \frac{\int_{0}^{{\cal T}} \frac{1}{2} m  \| v \|^2 \,dt}{\cal N},
        \end{equation}
    where $\cal T \in \nR{+}$ is the task duration, $m = 4.82~kg$ is the OMAV mass and $\| v \|$ is its velocity norm. It evaluates the operator efficiency consumption at using the aerial vehicle. Lower is better.
    \item 
    The perceived workload $\mathcal{W}$ for each experimental condition. It is measured using the NASA-TLX questionnaire~\cite{nasa-tlx}, assessing the workload experienced by participants during the task and includes six dimensions:
    \begin{enumerate*}
        \item \ac{MD} is the perceived cognitive load,
        \item \ac{PD} is the perceived physical effort,
        \item \ac{TD} is the perceived time pressure,
        \item \ac{EF} is the perceived level of exertion required,
        \item \ac{PE} is the perceived success in achieving the task goal, and
        \item \ac{FR} is the perceived level of stress during the task.
    \end{enumerate*}
\end{itemize}

\subsection{Data Collection and Analysis} \label{sec:data-collection}
\begin{figure}[tp]
  \centering
  \includegraphics[width=\linewidth]{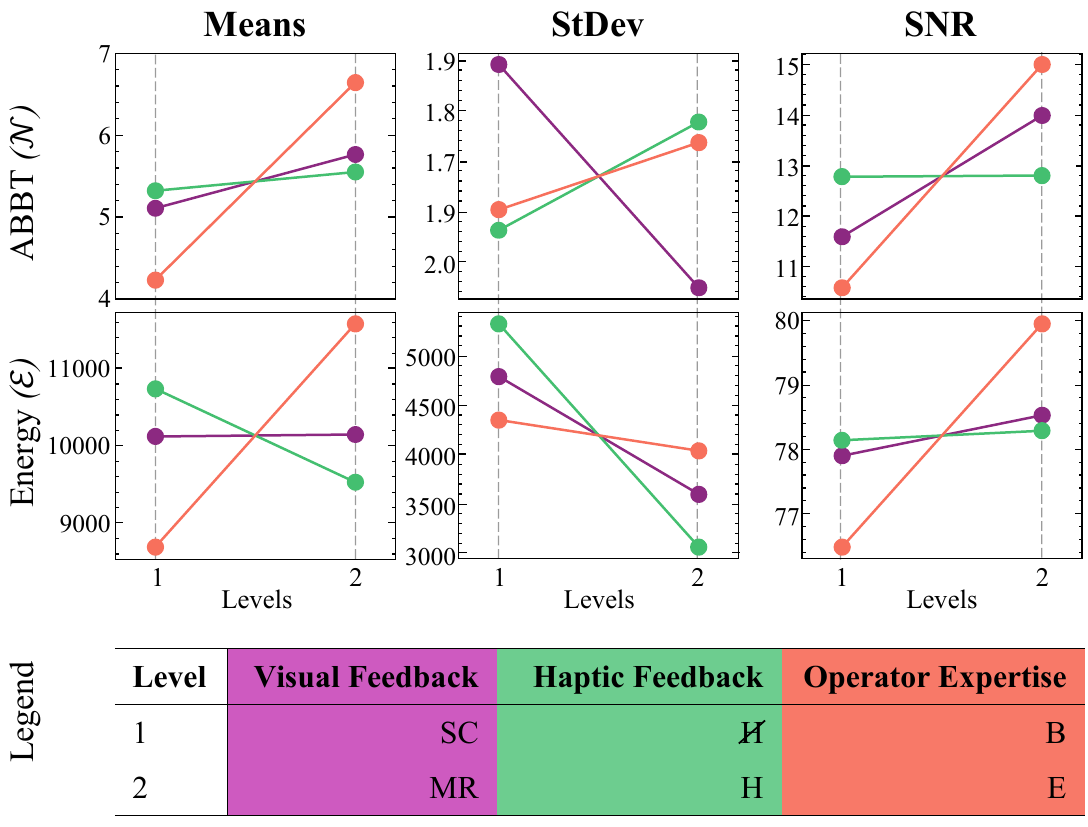}
  \caption{Main effects plots showing the individual influence of factors on response characteristics (i.e., Means, Standard Deviations and \ac{SNR}). The main effect shows the direction and magnitude of the effect that each factor has on the response, highlighting the identification of the optimal levels for those factors.}
  \label{fig:main-effect}
  \vspace{-0.5cm}
\end{figure}

Taguchi design allows to identify the optimal combination levels (see Sec.~\ref{sec:doe}) for each factor that maximize/minimize the selected metrics. Factor rank indicates its relative importance, with $Rank = 1$ signifying the highest impact on the result. The main effects plots of Taguchi are presented in Fig.~\ref{fig:main-effect}. In our case, we aim to maximize the mean number of blocks transferred ($\cal N$) among the experiments, while minimizing the mean energy spent ($\cal E$). We also want to minimize the standard deviation (StDev) to reduce variability in results. The \ac{SNR} evaluates the system's quality by comparing the desired signal with the unwanted variation in the output. A higher \ac{SNR} indicating better performance by minimizing the influence of noise on the experimental output.

To apply \ac{ANOVA}, data have to comply with the Shapiro-Wilk normality test \cite{anova-regression}. The mixed \ac{ANOVA} has been conducted with a significance level set at $0.05$, unless stated otherwise. All data passed Shapiro-Wilk normality test, except for SC-\cancel{H}-B in metric $\cal E$ with $p$-values $<0.010$. Post hoc analysis yielded significant p-values, which are presented in Tab.~\ref{tab:stat}. Additionally, significant p-values from the consideration of simple main effects are also reported in Fig.~\ref{fig:ANOVA}. Post hoc tests are used to perform pairwise comparison between the different setups. For the comparison of all the metrics, we conducted mixed \ac{ANOVA} tests. 

If a statistically significant interaction between variables is observed, conducting an analysis of the simple main effects becomes necessary \cite{anova-regression}. Tukey's method is used to identify confidence intervals for pairwise differences between factor level means.
By using the confidence intervals of the differences we determine whether the differences are practically significant, and examine their confidence intervals.
Using the grouping table in Tab.~\ref{tab:stat}, we can quickly identify statistically significant mean differences in groups not sharing a common letter. Results are presented in Sec.~\ref{sec:discussion}.

After the experiment, participants were asked to complete a \ac{NASA-TLX} questionnaire. The aim is to understand how workload subscales are perceived with the different setups. User filled first a pairwise comparison of the different subscales (\ac{MD}, \ac{PD}, \ac{TD}, \ac{EF}, \ac{PE}, \ac{FR}) to associate a rate between 0 and 5. Then, participants were asked to rate subscales from 0 to 100. A score of 0 represents "no workload", and a score of 100 represents "extreme workload". 
The adjusted workload coefficient is the product of the pairwise with the subscales. Because we found non-normal data from \ac{NASA-TLX}, a Mood's Median nonparametric test has been run \cite{nonparametric-mood}.
\vspace{-0.5cm}
\section{RESULT AND DISCUSSION}\label{sec:discussion}

\begin{figure}[tp]
  \centering
  \includegraphics[width=\linewidth]{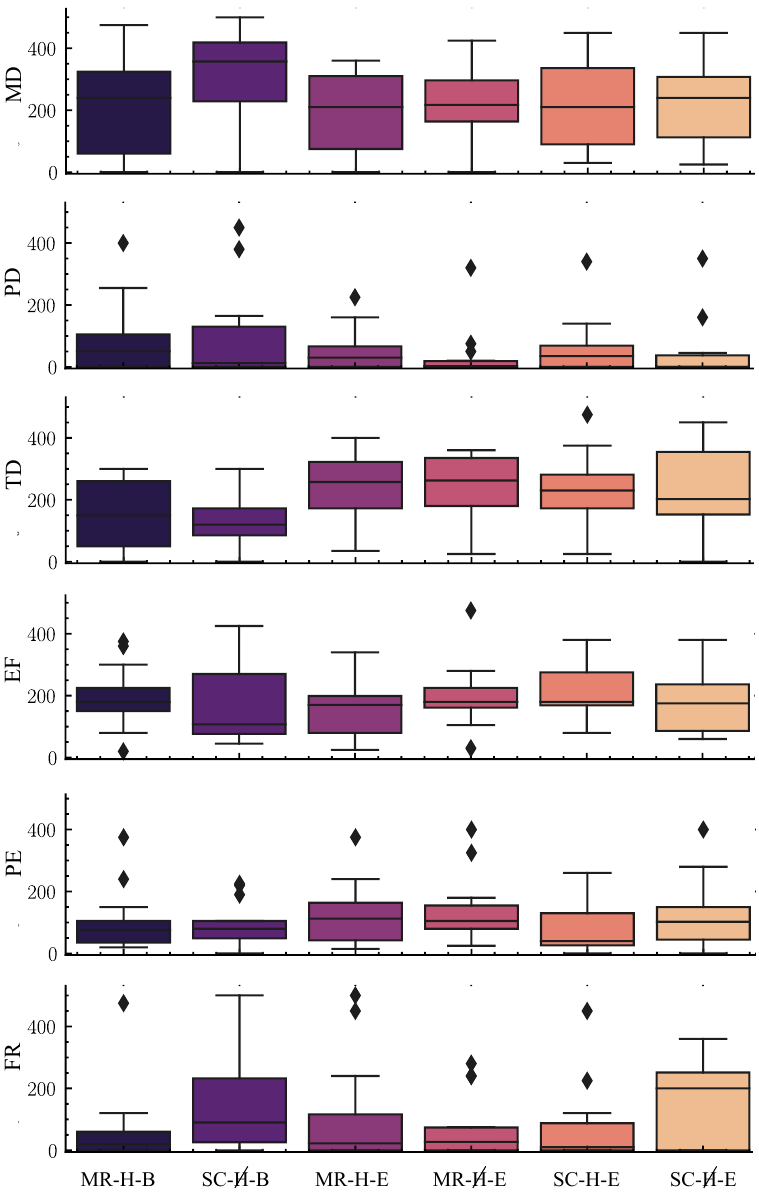}
  \caption{Distributions of the adjusted workload subscales}
  \label{fig:NASA-TLX}
  \vspace{-0.5cm}
\end{figure}

Main effects plots from the Taguchi design are shown in Fig.~\ref{fig:main-effect}. Data distributions of the study are plotted in Fig.~\ref{fig:ANOVA}, while distribution from the \ac{NASA-TLX} is in Fig.~\ref{fig:NASA-TLX}. The explanation and interpretation of the results follow.


\begin{center}
    \begin{figure}[tp]
        \subfloat[Quantity of blocks transferred]{\includegraphics[width=\linewidth]{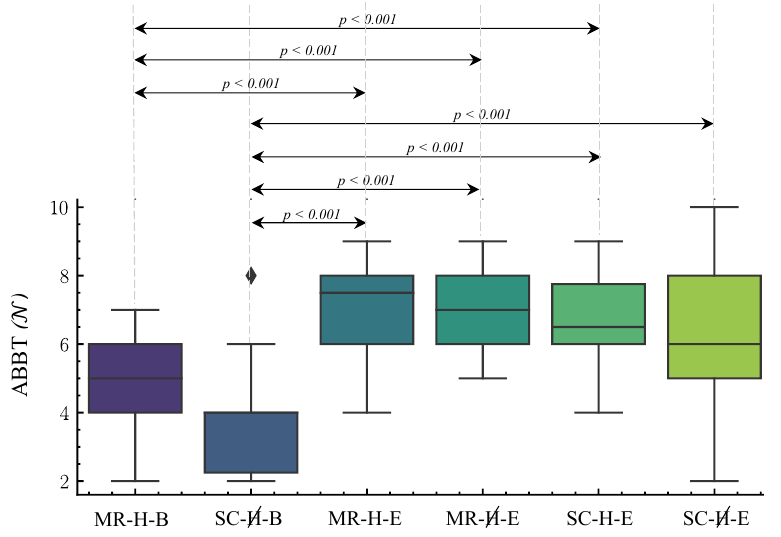}
        \label{fig:ANOVA-ABBT}}\\
        \subfloat[Robot energy consumed per block transfer]{\includegraphics[width=\linewidth]{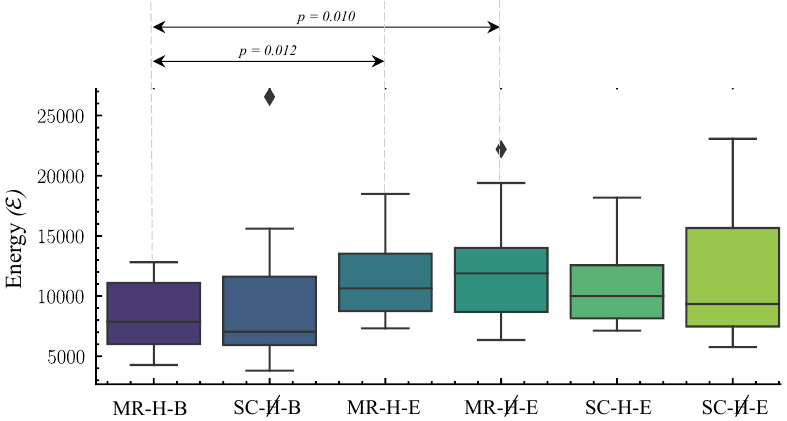}
        \label{fig:ANOVA-ratioEA}}\\
        \caption{Human-subjects study results from \ac{ANOVA}. Boxplot distribution of metrics for the different setup configurations. For (\ref{fig:ANOVA-ABBT}) the quantity of block transferred $\cal N$, and (\ref{fig:ANOVA-ratioEA}) the energy consumed $\cal E$ per block transferred.}
        \label{fig:ANOVA}
    \end{figure}
    \vspace{-0.8cm}
\end{center}

\begin{table}[t]
\def\arraystretch{1.05}
\small
\centering
 \caption{Experimental evaluation summary}
 \label{tab:stat} \vspace*{-2mm}
  \begin{tabular}{p{1.3cm} p{0.0cm} p{6.2cm}}
  \toprule
   \textbf{Subjects} & \multicolumn{2}{l}{28 (14 beginners, 14 experienced)}\\ 
     \midrule
      \textbf{Task} & \multicolumn{2}{l}{\begin{minipage}[c]{6.2cm} Teleoperation of the aerial manipulator to move blocks from an initial position to a target area.\end{minipage} }
      \\ 
\midrule
      \textbf{Factors} & \multicolumn{2}{l}{\underline{Operator Expertise}}\\
      & & B (beginner), E (experienced operator)\\
      & \multicolumn{2}{l}{\underline{Display}}\\ 
      & & SC (standard 2D screen), MR (MR 3D display)\\
      & \multicolumn{2}{l}{\underline{Feedback}}\\ 
      & & H (haptic feedback), \cancel{H} (no haptic feedback)\\
      \bottomrule
  \end{tabular}
 \vspace*{0.3cm}
\vspace*{-2mm}
\setlength{\tabcolsep}{0.5em} 
 \begin{tabular}{p{0.0cm}p{0.0cm}p{1.95cm}p{1.25cm}p{1.98cm}p{1.25cm}}
  \toprule
   \multicolumn{6}{l}{\textbf{Statistical analysis} (only significant $p$ values are shown)}\\ 
\multicolumn{6}{l}{\small\underline{Blocks quantity transferred, $\cal N$}} \\
    &  \multicolumn{3}{l}{\small Simple main effect}\\
    &  & \scriptsize SC-\cancel{H}-B \tiny vs \scriptsize SC-H-E & \scriptsize $p<0.001$ & \scriptsize MR-H-B \tiny vs \scriptsize SC-H-E & \scriptsize $p<0.001$\\
    &  & \scriptsize SC-\cancel{H}-E \tiny vs \scriptsize SC-\cancel{H}-B & \scriptsize $p<0.001$ & \scriptsize MR-H-E \tiny vs \scriptsize SC-\cancel{H}-B & \scriptsize $p<0.001$\\
    &  & \scriptsize MR-\cancel{H}-E \tiny vs \scriptsize SC-\cancel{H}-B & \scriptsize $p<0.001$ & \scriptsize MR-H-B \tiny vs \scriptsize SC-\cancel{H}-E & \scriptsize $p=0.021$\\
    &  & \scriptsize MR-H-E\hspace{0.05em}\tiny vs\hspace{0.05em}\scriptsize MR-H-B &\scriptsize $p<0.001$ & \scriptsize MR-\cancel{H}-E\tiny vs \scriptsize MR-H-B & \scriptsize $p<0.001$\\
    &  \multicolumn{3}{l}{\small Grouping information}\\
    &  & \footnotesize MR-H-E & \footnotesize A &  & \\
    &  & \footnotesize MR-\cancel{H}-E & \footnotesize A &   & \\
    &  & \footnotesize SC-H-E & \footnotesize A &  & \\
    &  & \footnotesize SC-\cancel{H}-E & \footnotesize A & \footnotesize B & \\
    &  & \footnotesize MR-H-B &  & \footnotesize B & \footnotesize C \\
    &  & \footnotesize SC-\cancel{H}-B &  &  & \footnotesize C \\
\multicolumn{6}{l}{\small\underline{Energy consumed per move, $\cal E$}} \\
    &  \multicolumn{3}{l}{\small Simple main effect of feedback}\\
    &  & \scriptsize MR-\cancel{H}-E vs MR-H-B & \footnotesize $p = 0.010$ & \scriptsize MR-H-E vs MR-H-B & \footnotesize $p = 0.012$\\
\multicolumn{6}{l}{\small\underline{Workload subscales, $\cal W$}} \\
    &  \multicolumn{3}{l}{\small Nonparametric test for main effect}\\
    &  & \small TD & \footnotesize $p = 0.013$ & & \\
  \end{tabular}  
\vspace*{-0.3cm}
\end{table}

\subsection{Haptics and 3D Vision Improve Dexterity}
The use of haptics and 3D MR vision were independently beneficial to the user. From Fig.~\ref{fig:main-effect} we can notice means of MR visual and haptic feedback improving user dexterity with respectively ($Delta = 0.658$, $Rank = 2$) and ($Delta = 0.230$, $Rank = 3$). Visual feedback reduced standard deviation and increased \ac{SNR} ($Delta = 2.41$, $Rank = 2$), with more blocks $\cal N$ moved during the \ac{ABBT}. Haptic feedback improved the overall user dexterity, but contrary to visual feedback, it comes with a slight dispersion, and therefore a smaller performance improvement than with MR. A possible explanation is that users interpret force feedback differently and, if its meaning is not well understood, it can be perceived as a perturbation instead of a clear force information, especially for some beginners that reported it.

From the grouping information (Tab.~\ref{tab:stat}) we confirm the previous results and add precision. The second column (referred to with letter B) shows significant difference with the groups containing MR and/or H, and the group using SC-\cancel{H}. Thus, the use of haptics and/or MR vision robustifies the operator's dexterity. A beginner with the setup using MR-H, compared to an experienced operator with SC-\cancel{H} (MR-H-B vs. SC-\cancel{H}-E) are not significantly different in dexterous aerial manipulation. 
Furthermore, with the number of block displaced (Fig.~\ref{fig:ANOVA-ratioEA}), contrary to energy efficiency (Fig.~\ref{fig:ANOVA-ABBT}) where MR vision seems to be more important, a power-saving flight looks to take more advantage of haptic feedback. In other words, haptics allows longer flight time.

\subsection{Training and Transfer Learning}
The most important factor at improving blocks quantity transferred $\cal N$ in \ac{ABBT} is the operator experience. This can be seen from the means ($Delta = 2.413$, $Rank = 1$) in Fig.~\ref{fig:main-effect}. User experience also improves repeatability of the result as one can see from the \ac{SNR} ($Delta = 4.442$, $Rank = 1$). Experienced operators look to better exploit the system, which is confirmed by looking at the energy consumption ($Delta = 2892$, $Rank = 1$). This result can be explained by the fact that beginners fly more smoothly, as they don't know the limits of the system. On the other hand, experienced pilots are able to accelerate faster.

By running two-way ANOVA, Fig.~\ref{fig:ANOVA-ABBT} reveals a significant difference for the operator expertise with $p$-value $< 0.001$ ($F=46.70$). Post hoc Tukey test shows that simple main effects of SC-\cancel{H}-B with each E configuration has a statistical significant difference. Likewise, the MR-H-B configuration has statistically significant difference with MR-H-E, 3D-\cancel{H}-E and SC-H-E. This confirms result coming from Taguchi, but specify that only operator experience is significant. Thus, an experienced operator takes advantage of the 3D \ac{MR} vision and haptics offered by the setup.
Then, we want to minimize the energy consumption $\cal E$. Figure~\ref{fig:main-effect} shows that operator expertise ($Delta = 3.48$, $Rank = 1$) has still the largest effect. Running the two-way ANOVA (see Fig.~\ref{fig:ANOVA-ratioEA}) reveals a significant difference for the levels of expertise of the operator with a $p$-value $= 0.006$ ($F=7.84$). Post hoc Tukey Method shows statistically significant difference of MR-H-E vs. MR-H-B ($F=7.25$, $p=0.012$) and MR-\cancel{H}-E vs. MR-H-B ($F=7.74$, $p=0.010$). As a result, an experienced operator will perform better at the task, but will consume more energy with the robot, that in turn results in a reduction in flight time.

We can also notice from the \ac{ABBT} a transfer learning, from the training with haptics and 3D \ac{MR} vision, to the setup without haptics and 2D screen. Post hoc Tukey Method shows statistically significant difference with SC-\cancel{H}-B vs. SC-\cancel{H}-E ($F=24.13$, $p < 0.001$). This means that the experienced group transferred their capacity obtained with our augmented setup, to the one without haptics and a conventional screen.

\subsection{Limited Influence on Workload}
For the \ac{NASA-TLX}, because none of the subscales workload data passed Shapiro-Wilk normality test, we ran Mood's Median nonparametric test.
The results are plotted in Fig.~\ref{fig:NASA-TLX}. Only \ac{TD} accepted the null hypothesis, \textit{population medians are all equal}, with ($\chi^2 = 14.53$, $p=0.013$). The different setups seem to not influence the workload perception of the user. A significant difference has been noticed for \ac{TD}. Experienced operators are more involved in carrying out the task, and are therefore more impacted by time spent on the task, than beginners. On the other hand, when using the 2D screen (Fig.~\ref{fig:setup-a}), participants used the shadows to get the relative position of the robot with respect to the payload. Participants also reported that using a standard 2D screen requires more attention to fly. This tendency can be seen on \ac{MD} of Fig.~\ref{fig:NASA-TLX} without being significant.
We observed several interesting tendencies that, however, are not statistically significant due to data dispersion. The tendency for \ac{PD} to be more felt by the beginners than experienced operators can be attributed to the latter's habit of using the device. Additionally, the observation that H-E for \ac{PD} is greater than \cancel{H}-E can be explained by the additional force provided by the joystick. The perceived \ac{PE} of beginners is lower than that of experienced, who was more aware of their capabilities due to the training. The \ac{EF} was slightly less for MR-H-E, which was the training factor configuration. Furthermore, some participants verbalized that using haptics reduced \ac{FR}. Conversely, we observed a tendency for the standard setup with SC-\cancel{H} to increase \ac{FR} for both beginners and experienced.


\subsection{Limitations}
To evaluate human dexterity we chose to adapt the \ac{BBT} for aerial manipulation, because it has already been used successfully in virtual environments~\cite{vbbt}. However there exist other reliable tests such as the Minnesota Manual Dexterity Test (MMDT) and the Purdue Pegboard (PP). These two tasks, which involve peg-in-hole operation, could also benefit from haptic bilateral feedback. Nevertheless, there is a significant correlation of MMDT and PP with the \ac{BBT}~\cite{dexterity-tasks}. Thus, evaluating the operator's dexterity in accurately picking up the box during dynamic movement is a reasonable approach.

From a virtual rendering point of view, an interesting point to highlight is that literature has shown a difference in how humans perceive objects in \ac{VR} vs. \ac{MR}~\cite{ar-softer, ahn2019size}. Such a difference, although limited to a few sensory features such as stiffness and shape, might have played a role in the outcome of our experiment. Another point to highlight is the reliance on a motion capture system for localization. However, this was not a central focus of our research and does not affect the validity of our findings. The generalization of the system to work in other environments with on board rendering~\cite{virtual-rendering} can be addressed with widely available and effective method, like visual dynamic object-aware SLAM system~\cite{vdo-slam}.





\section{CONCLUSION}
This work studies the effect of 3D \ac{MR} vision and haptics on operator performance in the context of aerial tele-manipulation. A multimodal bilateral teleoperation system based on \ac{MR} vision and haptic force feedback has been developed for aerial manipulation tasks. 
Experimental \ac{APhI} tasks demonstrate sub-centimeter-level accuracy and precise force application when teleoperating aerial manipulators using the proposed system. 
Results of a user study enrolling $28$ participants confirm quantitatively that both interaction feedback and MR-based visualization indeed improve user dexterity and flight efficiency. 
Nevertheless, the experience of the operator is still the most important factor.


Since the training of the experienced group only happened with 3D \ac{MR} vision and haptics, future direction would question the learning rate at using or not the system. In robot assisted surgery, haptic feedback has been proved crucial in early psychomotor skill acquisition \cite{learning-surgery-robot}. Its potential generalisation to aerial manipulation has to be explored. The combination with the transfer learning we observed would reduce the training duration of operators.

\bibliographystyle{IEEEtran}
\bibliography{bibliography.bib}


 

\begin{IEEEbiography}
[{\includegraphics[width=1in,height=1.25in,clip,keepaspectratio]{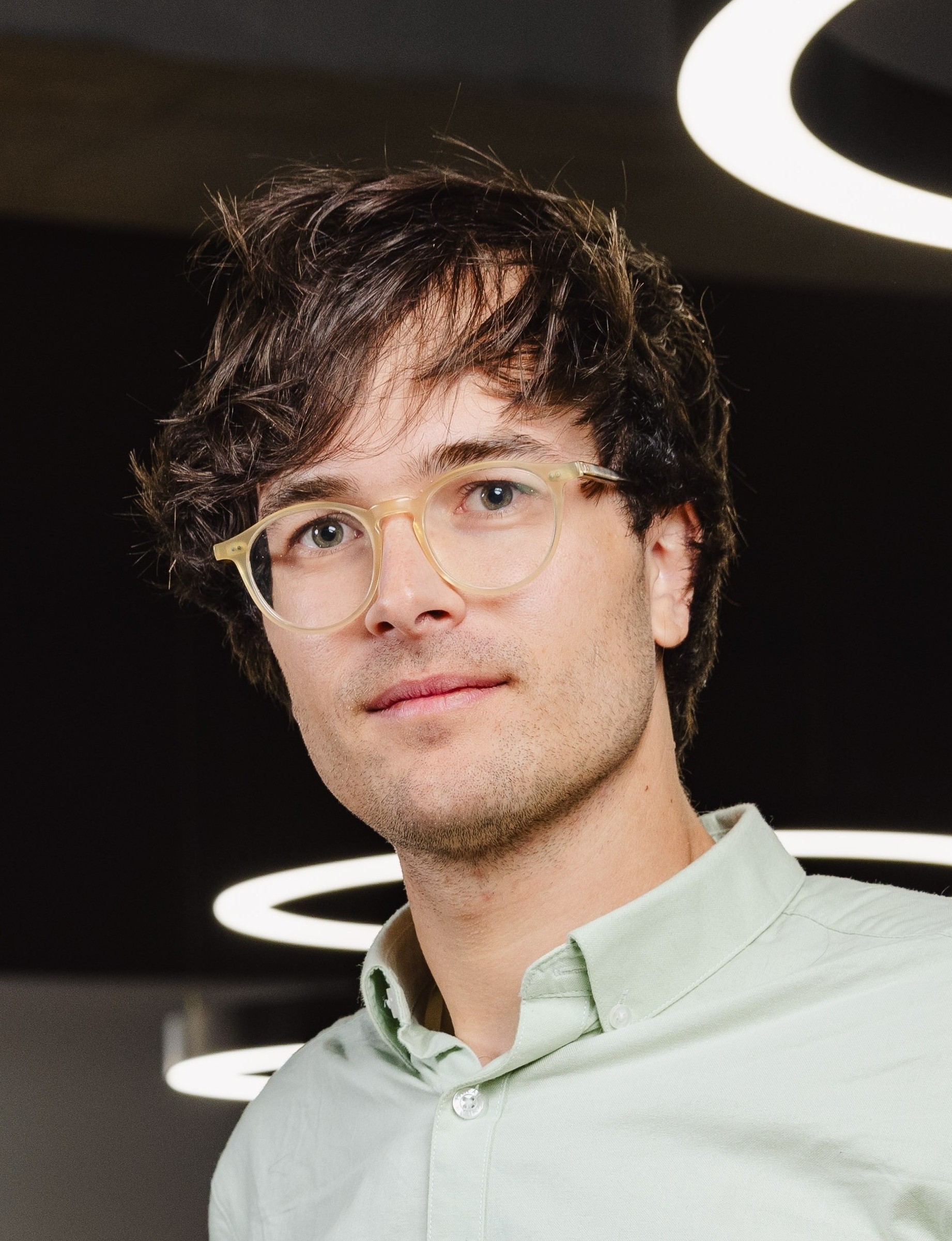}}]{Julien Mellet}
received the engineering degree from INSA Centre Val de Loire, France, in industrial system engineering, and the M.Sc. degree from NPU, Xi'an, China, in navigation, guidance and control in 2018. He is currently working toward the Ph.D. degree on aerial manipulation at the PRISMA Lab, University of Naples Federico II, Italy with a Marie Sklodowska-Curie scholarship. His research interests include the design and control of multi-robot systems for aerial manipulation with haptic feedback in the context of teleoperation.
\end{IEEEbiography}

\begin{IEEEbiography}
[{\includegraphics[width=1in,height=1.25in,clip,keepaspectratio]{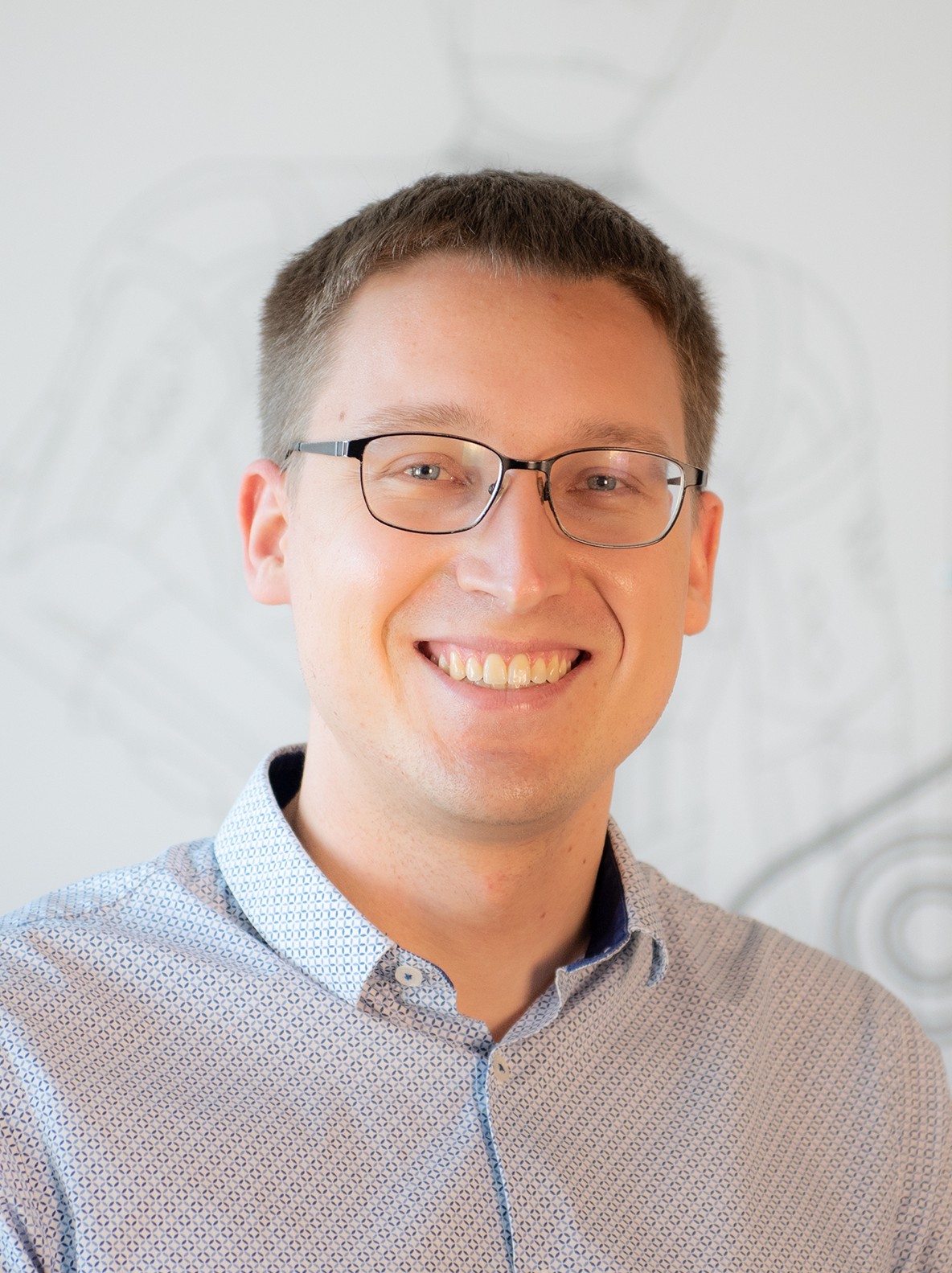}}]{Mike Allenspach} received his Master degree in
Robotics, Systems and Control in 2020 from ETH Zurich,
Switzerland. He is currently pursuing a Ph.D. degree at
the Autonomous Systems Lab at ETH Zurich. His research interests include planning and control for aerial manipulation, especially the integration of human-robot interaction.

\end{IEEEbiography}

\begin{IEEEbiography}
[{\includegraphics[width=1in,height=1.25in,clip,keepaspectratio]{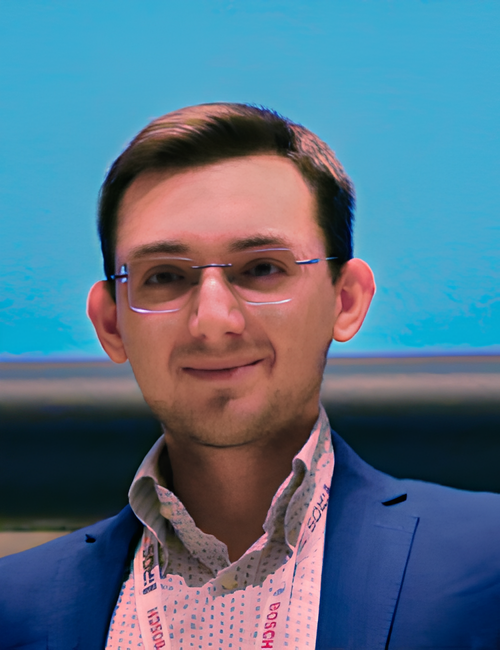}}]{Eugenio Cuniato}
(Graduate Student Member, IEEE) received his Master degree in Automation Engineering from the University of Naples Federico II, Naples, Italy, in 2021. He is currently pursuing a Ph.D. degree at the Autonomous Systems Lab, ETH Zurich, Zurich, Switzerland with a Marie Sklodowska-Curie scholarship. His research focuses on control of aerial manipulators for aerial physical interaction. He received the Best Paper Award on Safety, Security, and Rescue Robotics at the IEEE IROS 2022 conference and the IEEE Robotics and Automation Letters Distinguished Service Award in 2024.
\end{IEEEbiography}

\begin{IEEEbiography}
[{\includegraphics[width=1in,height=1.25in,clip,keepaspectratio]{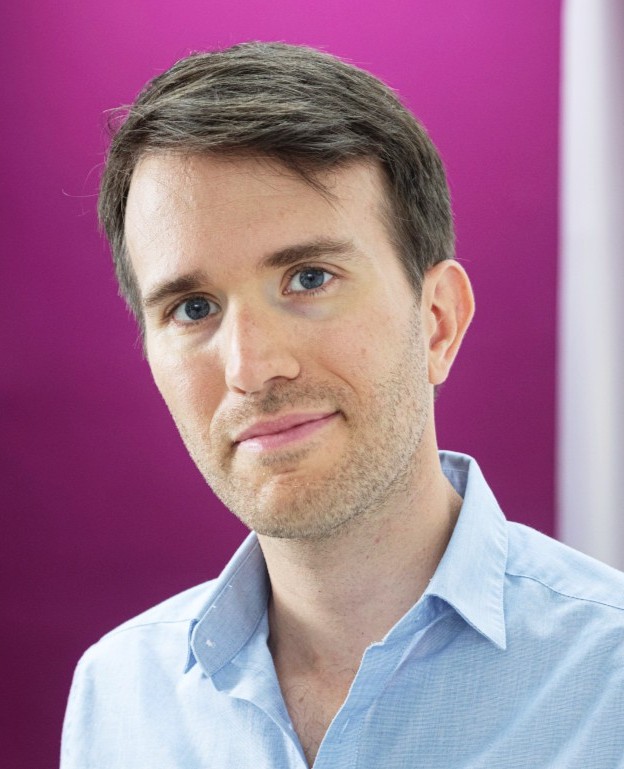}}]{Claudio Pacchierotti}
(Senior Member, IEEE) received the Ph.D. degree with the University of Siena, Siena, Italy, in 2014. Since 2016, he has been a tenured researcher with CNRS-IRISA, Rennes, France. He was previously a Postdoctoral Researcher with the Italian Institute of Technology, Genova, Italy. He was Visiting Researcher of the Penn Haptics Group, University of Pennsylvania, USA, in 2014, the Dept. Innovation in Mechanics and Management, University of Padua, Italy, in 2013,  the Institute for Biomedical Technology and Technical Medicine (MIRA), University of Twente, The Netherland, in 2014, and the Dept. Computer, Control and Management Engineering, Sapienza University of Rome, Italy, in 2022. He was the recipient of the 2014 EuroHaptics Best PhD Thesis Award and the 2022 CNRS Bronze Medal. He is Senior Chair of the IEEE Technical Committee on Haptics, Co-Chair of the IEEE Technical Committee on Telerobotics, and Secretary of the Eurohaptics Society.

\end{IEEEbiography}

\begin{IEEEbiography}
[{\includegraphics[width=1in,height=1.25in,clip,keepaspectratio]{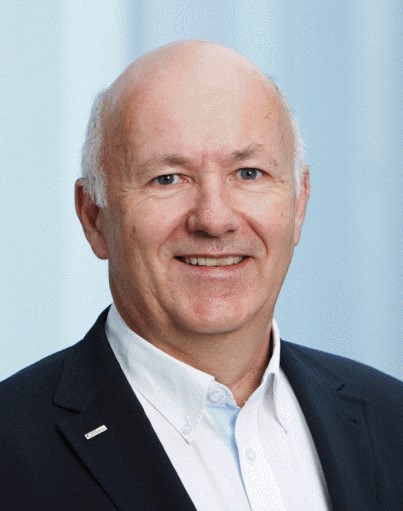}}]{Roland Siegwart}
(Fellow, IEEE) is a Professor of autonomous mobile robots with ETH Zurich, Zurich, Switzerland; the Founding Co-Director of the Technology Transfer Center, Wyss Zurich, Zurich; and a Board Member of multiple high-tech companies. From 1996 to 2006, he was a Professor with EPFL Lausanne. He has held a visiting positions with Stanford University and NASA Ames. He was the Vice President of ETH Zurich, from 2010 to 2014. His research interests include the design, control, and navigation of flying and wheeled and walking robots operating in complex and highly dynamical environments. Prof. Siegwart was the recipient of the IEEE RAS Pioneer Award and the IEEE RAS Inaba Technical Award. He is among the most cited scientist in robots world-wide, a Co-Founder of more than half a dozen spin-off companies, and a strong promoter of innovation and entrepreneurship in Switzerland.
\end{IEEEbiography}

\begin{IEEEbiography}
[{\includegraphics[width=1in,height=1.25in,clip,keepaspectratio]{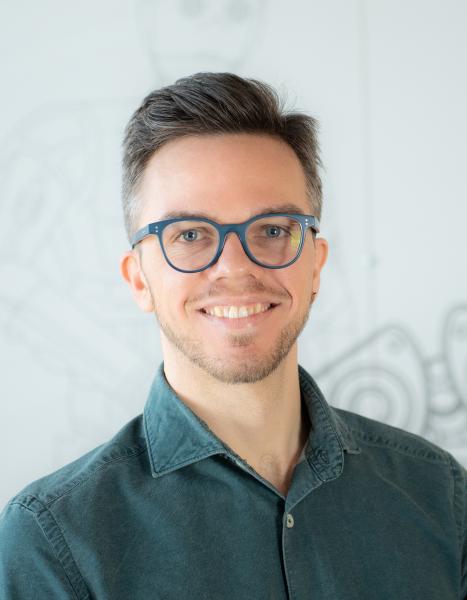}}]{Marco Tognon}
(Member, IEEE) received the M.Sc. degree in automation engineering from the University of Padua, Padua, Italy, in 2014, with a master thesis carried out at the Max Planck Institute for Biological Cybernetics, Germany, and the Ph.D. degree in robotics from the INSA Toulouse, Toulouse, France, in 2018, developing his thesis at LAAS-CNRS, Toulouse. Since November 2022, he has been a Tenured Researcher with the Rainbow Team, INRIA Rennes, Rennes, France. Before that, he was a Postdoctoral Researcher with the Autonomous System Lab, ETH Zurich, Switzerland. From 2018 to 2020, he was a Postdoctoral Researcher with LAAS-CNRS. His current research interests include robotics, aerial physical interaction, multirobot systems, and human—robot physical interaction
\end{IEEEbiography}

\vfill

\end{document}

%% file: symbol_definitions.tex
\usepackage{acro}
\usepackage{bm}
\usepackage{mathtools}
\usepackage{xcolor}

\DeclareAcronym{ASL}{short = ASL, long = Autonomous Systems Lab}
\DeclareAcronym{OMAV}{short = OMAV, long = Omnidirectional Micro Aerial Vehicle}
\DeclareAcronym{MAV}{short = MAV, long = Micro Aerial Vehicle}
\DeclareAcronym{DoF}{short = DoF, long = degrees of freedom}
\DeclareAcronym{PBC}{short = PBC, long = passivity-based control}
\DeclareAcronym{PH}{short = PH, long = Port-Hamiltonian}
\DeclareAcronym{NDT}{short = NDT, long = non-destructive testing}
\DeclareAcronym{DOE}{short = DOE, long = design of experiments}
\DeclareAcronym{PEMS}{short = PEMS, long = Power and Energy Monitoring System}
\DeclareAcronym{WTC}{short = WTC, long = wrench tracking controller}
\DeclareAcronym{PTC}{short = PTC, long = pose tracking controller}
\DeclareAcronym{MBE}{short = MBE, long = momentum-based wrench estimator}
\DeclareAcronym{ASIC}{short = ASIC, long = Axis-Selective Impedance Control}
\DeclareAcronym{MPC}{short = MPC, long = Model Predictive Control}
\DeclareAcronym{MPPI}{short = MPPI, long = Model Predictive Path Integral}
\DeclareAcronym{APhI}{short = APhI, long = Aerial Physical Interaction}
\DeclareAcronym{LLE}{short = LLE, long = Largest Lyapunov Exponent}
\DeclareAcronym{ICBF}{short = ICBF, long = Integral Control Barrier Function}
\DeclareAcronym{CBF}{short = CBF, long = Control Barrier Function}
\DeclareAcronym{COM}{short = CoM, long = Center of Mass}
\DeclareAcronym{AM}{short = AM, long = Aerial Manipulator}
\DeclareAcronym{MR}{short = MR, long = Mixed Reality}
\DeclareAcronym{AR}{short = AR, long = Augmented Reality}
\DeclareAcronym{VR}{short = VR, long = Virtual Reality}
\DeclareAcronym{HRI}{short = HRI, long = Human-Robot Interaction}
\DeclareAcronym{RL}{short = RL, long = Reinforcement Learning}
\DeclareAcronym{PPO}{short = PPO, long = Proximal Policy Optimization}

\DeclareAcronym{NASA-TLX}{short = NASA-TLX, long = NASA Task Load Index}
\DeclareAcronym{MD}{short = MD, long = mental demand}
\DeclareAcronym{PD}{short = PD, long = physical demand}
\DeclareAcronym{TD}{short = TD, long = temporal demand}
\DeclareAcronym{EF}{short = EF, long = effort}
\DeclareAcronym{PE}{short = PE, long = performance}
\DeclareAcronym{FR}{short = FR, long = frustration}
\DeclareAcronym{SNR}{short = SNR, long = signal-to-noise ratio}
\DeclareAcronym{ANOVA}{short = ANOVA, long = Analyse of Variance}
\DeclareAcronym{LCD}{short = LCD, long = liquid crystal display}
\DeclareAcronym{FT}{short = F/T, long = force and torque, short-indefinite = an, long-indefinite = a}
\DeclareAcronym{BBT}{short = BBT, long = Box and Block Test}
\DeclareAcronym{ABBT}{short = ABBT, long = Aerial Box and Block Test}
\DeclareAcronym{MOCAP}{short = MOCAP, long = Motion Tracking System}

\renewcommand{\vec}[1]{\bm{#1}}		
\newcommand{\matr}[1]{\bm{#1}}		
\newcommand{\nR}[1]{\mathbb{R}^{#1}}		
\newcommand{\nN}[1]{\mathbb{N}^{#1}}		
\newcommand{\SO}[1]{\mathsf{SO}(#1)}		
\newcommand{\upperRomannumeral}[1]{\uppercase\expandafter{\romannumeral#1}}	

\newcommand{\transpose}{^\top}

\renewcommand{\frame}[1]{\mathcal{F}_{#1}}		

\newcommand{\origin}{O}						
\newcommand{\vX}{\vec{x}}					

\newcommand{\vY}{\vec{y}}					
\newcommand{\vZ}{\vec{z}}					
\newcommand{\pos}{\vec{p}_B}				
\newcommand{\posRef}{\vec{p}_{B,\text{ref}}}    
\newcommand{\vel}{\vec{v}_B}				
\newcommand{\velRef}{\vec{v}_{B,\text{ref}}}	
\newcommand{\posH}{\vec{p}_H}				
\newcommand{\velH}{\vec{v}_H}				
\newcommand{\accB}{\dot{\vec{v}}_B^B}	

\newcommand{\velMax}{{v}_{max}}
\newcommand{\rotMat}{\matr{R}}				

\newcommand{\frameW}{\frame{W}}			
\newcommand{\frameB}{\frame{B}}			
\newcommand{\frameM}{\frame{M}}			
\newcommand{\frameH}{\frame{H}}			
\newcommand{\frameV}{\frame{V}}			
\newcommand{\originW}{\origin_W}		
\newcommand{\originH}{\origin_H}		
\newcommand{\originB}{\origin_B}		
\newcommand{\originM}{\origin_M}		
\newcommand{\originV}{\origin_V}		
\newcommand{\xW}{\vX_W}				
\newcommand{\yW}{\vY_W}				
\newcommand{\zW}{\vZ_W}				
\newcommand{\xB}{\vX_B}				
\newcommand{\yB}{\vY_B}				
\newcommand{\zB}{\vZ_B}				
\newcommand{\xM}{\vX_M}				
\newcommand{\yM}{\vY_M}				
\newcommand{\zM}{\vZ_M}				
\newcommand{\xH}{\vX_H}				
\newcommand{\yH}{\vY_H}				
\newcommand{\zH}{\vZ_H}				
\newcommand{\xV}{\vX_V}				
\newcommand{\yV}{\vY_V}				
\newcommand{\zV}{\vZ_V}				

\newcommand{\rotMatWB}{\rotMat_B^W}	
\newcommand{\rotMatWBRef}{\rotMat_{B,\text{ref}}^W}	
\newcommand{\rotMatMH}{\rotMat_H^M}	
\newcommand{\angVel}{\vec{\omega}_B}

\newcommand{\angVelB}{\dot{{\vec{\omega}}}_B^B}

\newcommand{\wrenchExt}{\wrench_\text{ext}}
\newcommand{\wrenchEst}{\hat{\wrench}_\text{ext}}
\newcommand{\wrenchTotal}{\wrench_\text{fb,total}}
\newcommand{\wrenchRec}{\wrench_\text{fb,rec}}
\newcommand{\wrenchFbExt}{\wrench_\text{fb,ext}}

\newcommand{\wrench}{\bm{\tau}}

